\documentclass{article}

\usepackage[T1]{fontenc}
\usepackage{amssymb,amsmath}
\usepackage{txfonts}
\usepackage{microtype}

\usepackage{graphicx}
\usepackage{subfigure} 

\usepackage{natbib}

\usepackage{algorithm}
\usepackage{algorithmic}

\usepackage{float}

\usepackage{hyperref}
\usepackage{url}
\usepackage{courier}

\usepackage{mlp2017}
\DeclareMathOperator{\sigmoid}{sigmoid}

\DeclareMathOperator{\relu}{relu}
\DeclareMathOperator{\lrelu}{lrelu}
\DeclareMathOperator{\elu}{elu}
\DeclareMathOperator{\selu}{selu}

\def\studentNumber{Giovanni Alcantara}

\begin{document} 

\twocolumn[
\mlptitle{Empirical analysis of non-linear activation functions for Deep Neural Networks in classification tasks}

\centerline{\textbf{\studentNumber}}
\centerline{Department of Computer Science}
\centerline{University of Edinburgh}
\centerline{University of Texas at Austin}
\centerline{\texttt{\{giovanni.alcantara\}@utexas.edu}}

\vskip 7mm
]

\begin{abstract}
We provide an overview of several non-linear activation functions in a neural network architecture that have proven successful in many machine learning applications. We conduct an empirical analysis on the effectiveness of using these function on the MNIST classification task, with the aim of clarifying which functions produce the best results overall. Based on this first set of results, we examine the effects of building deeper architectures with an increasing number of hidden layers. We also survey the impact of using, on the same task, different initialisation schemes for the weights of our neural network. Using these sets of experiments as a base, we conclude by providing a optimal neural network architecture that yields impressive results in accuracy on the MNIST classification task.
\end{abstract} 

\section{Introduction}
\label{sec:intro}
Building neural net architecture that work well in practice involves finding and tuning many optimal parameters and making several design decisions that make the issue not trivial.
One of such decision is the type of layer that compose the the neural net, particularly involving the type of activation functions that will be implemented as hidden layers.
The ability to model non-linearities in our data is at the foundation of using such activations function, which allow us to build more complex representations of our data.

Historically, non-linear activations functions like the logistic sigmoid functions or tanh functions, while successful with certain data distributions, have proven difficult to train, mostly due to their non-zero centered property and slope of the function \citep{DBLP:journals/corr/XuHL16}.
Many activations functions have been introduced in machine learning literature, with some surfacing as working well with many practical applications.

In section~\ref{sec:actfn}, we initially provide a theoretical overview of the activation functions used in our experiments.

We then provide, in section~\ref{sec:actexpts}, an empirical analysis and comparison of the performance of five such activation functions on a fixed task: the MNIST task of classifying hand-written digits \citep{20001258711}. We use a dataset of 60,000 data points, each being a 28x28 B\&W image of a handwritten representation of a digit. For each experiment, we split the dataset into a training set of 50,000 datapoints, with the remaining 10,000 used as validation set. We use a fixed number of epochs (100), regardless of weather our model converges earlier. We also use a batch size of 50.

Based on our results from this analysis, we follow in section~\ref{sec:deepmodelexp} with an exploration of the impact of the depth of our model for the same task, by ranging the number of hidden layers from 2 to 8.

Finally, in section~\ref{sec:weightinitexp} we analyse different initilisation schemas for our neural networks weights, and see how overall accuracies vary accordingly.

\section{Activation functions}
\label{sec:actfn}
This section covers a brief theoretical background on the activation functions taken into consideration in this work. We will use two activation functions as baseline models and four other functions as variations with potential optimizations to achieve better accuracy.
The baseline models are the sigmoid function and ReLU function (Rectified Linear Unit) \citep{DBLP:journals/corr/AroraBMM16}. We consider these two baseline models since they are commonly used in many machine learning applications, and particularly in the case of the ReLU function, have stemmed over the years, variations that account for different data distributions and have experimentally yielded better results. The other four activations functions are Leaky ReLU \citep{DBLP:journals/corr/XuWCL15}, ELU or Exponential Linear Units \citep{DBLP:journals/corr/ClevertUH15}, and SELU or Scaled Exponential Linear Units \citep{DBLP:journals/corr/KlambauerUMH17}.

\begin{figure}[H]
\vskip -3mm
\begin{center}
\centerline{\includegraphics[width=\columnwidth]{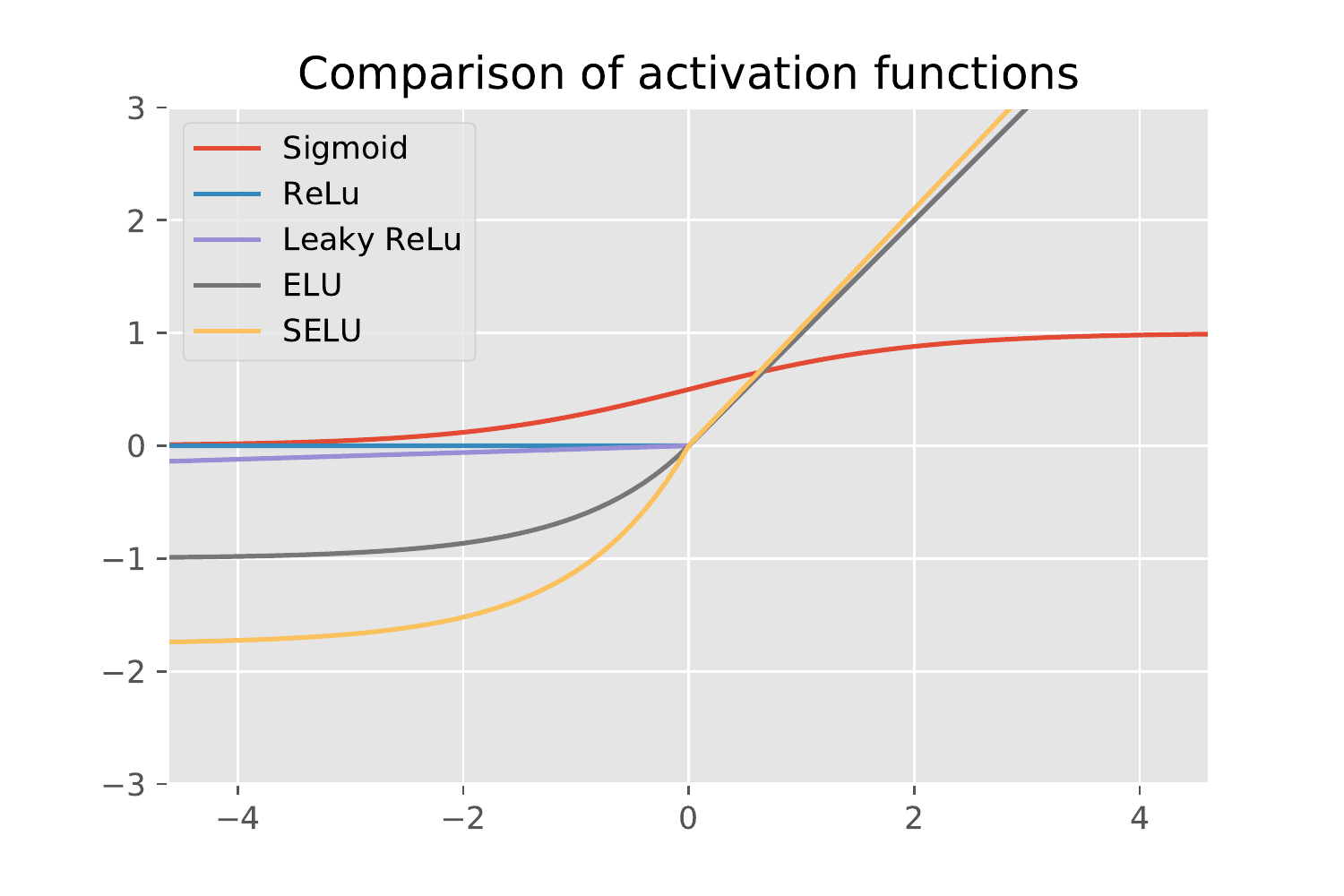}}
\caption{Plots of activation functions}
\label{fig:comparison_all}
\end{center}
\vskip -3mm
\end{figure}

Following are the definitions of the considered activations functions with their respective gradients, followed, in figure~\ref{fig:comparison_all}, by a their visualisation on the x-y axis.

\textbf{Sigmoid} (baseline model):
\begin{equation}
  \sigmoid(x) = \frac{1}{1+\exp{-x}}
\end{equation}

\begin{equation}
  \frac{d}{dx} \sigmoid(x) = \sigmoid(x)(1-\sigmoid(x))
\end{equation}

\textbf{ReLU} (baseline model):
\begin{equation}
  \relu(x) = \max(0, x)
\end{equation} 

\begin{equation}
  \frac{d}{dx} \relu(x) =
     \begin{cases} 
      0      & \quad \text{if } x \leq  0 \\
      1       & \quad \text{if } x > 0 .
    \end{cases} 
\end{equation}

\textbf{Leaky ReLU:}
\begin{equation}
  \lrelu(x) = 
  \begin{cases} 
      \alpha x      & \quad \text{if } x \leq  0 \\
      x       & \quad \text{if } x > 0 .
    \end{cases} 
\end{equation} 

\begin{equation}
  \frac{d}{dx} \lrelu(x) =
     \begin{cases} 
      \alpha      & \quad \text{if } x \leq  0 \\
      1       & \quad \text{if } x > 0 .
    \end{cases} 
\end{equation}

\textbf{ELU:}
\begin{equation}
  \elu(x) = 
  \begin{cases} 
      \alpha (\exp(x) - 1)      & \quad \text{if } x \leq  0 \\
      x       & \quad \text{if } x > 0 .
    \end{cases} 
\end{equation} 

\begin{equation}
  \frac{d}{dx} \elu(x) =
     \begin{cases} 
      \alpha \exp(x)      & \quad \text{if } x \leq  0 \\
      1       & \quad \text{if } x > 0 .
    \end{cases} 
\end{equation}

\textbf{SELU:}
\begin{equation}
  \selu(x) = \lambda
  \begin{cases} 
      \alpha (\exp(x) - 1)      & \quad \text{if } x \leq  0 \\
      x       & \quad \text{if } x > 0 .
    \end{cases} 
\end{equation} 

\begin{equation}
  \frac{d}{dx} \selu(x) = \lambda
     \begin{cases} 
      \alpha \exp(x)      & \quad \text{if } x \leq  0 \\
      1       & \quad \text{if } x > 0 .
    \end{cases} 
\end{equation}

\section{Experimental comparison of activation functions}
\label{sec:actexpts}

In this section we present the results and discussion of experiments comparing networks using the different activation functions on the MNIST task \citep{20001258711}.  We use 2 hidden layers with 100 hidden units per layer for these experiments. We report the learning curves (error vs epoch) for validation, and the validation set accuracies. 

We start the experimental comparison by analysing our baseline models, and how they perform with different learning rates. We use the following learning rate values in this section: 0.02, 0.05, 0.10. We fixed these values after experimentally defining lower and upper bound values that either make the model not converge during training, or that gets stuck in local minima.

\subsection{Sigmoid}

\begin{figure}[H]
\vskip -3mm
\begin{center}
\subfigure{
\includegraphics[width=38mm, height=30mm]{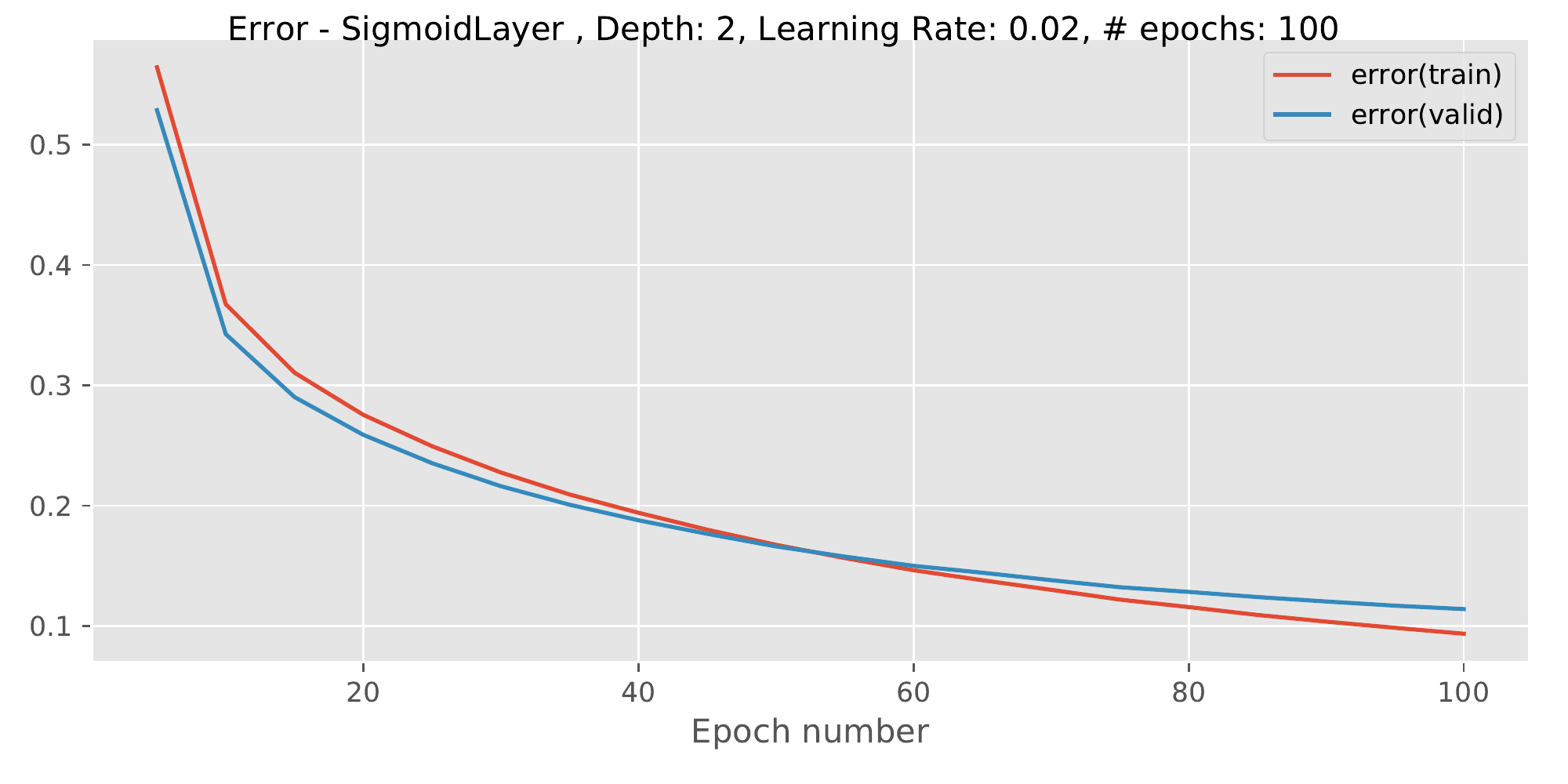}}
\subfigure{\includegraphics[width=38mm, height=30mm]{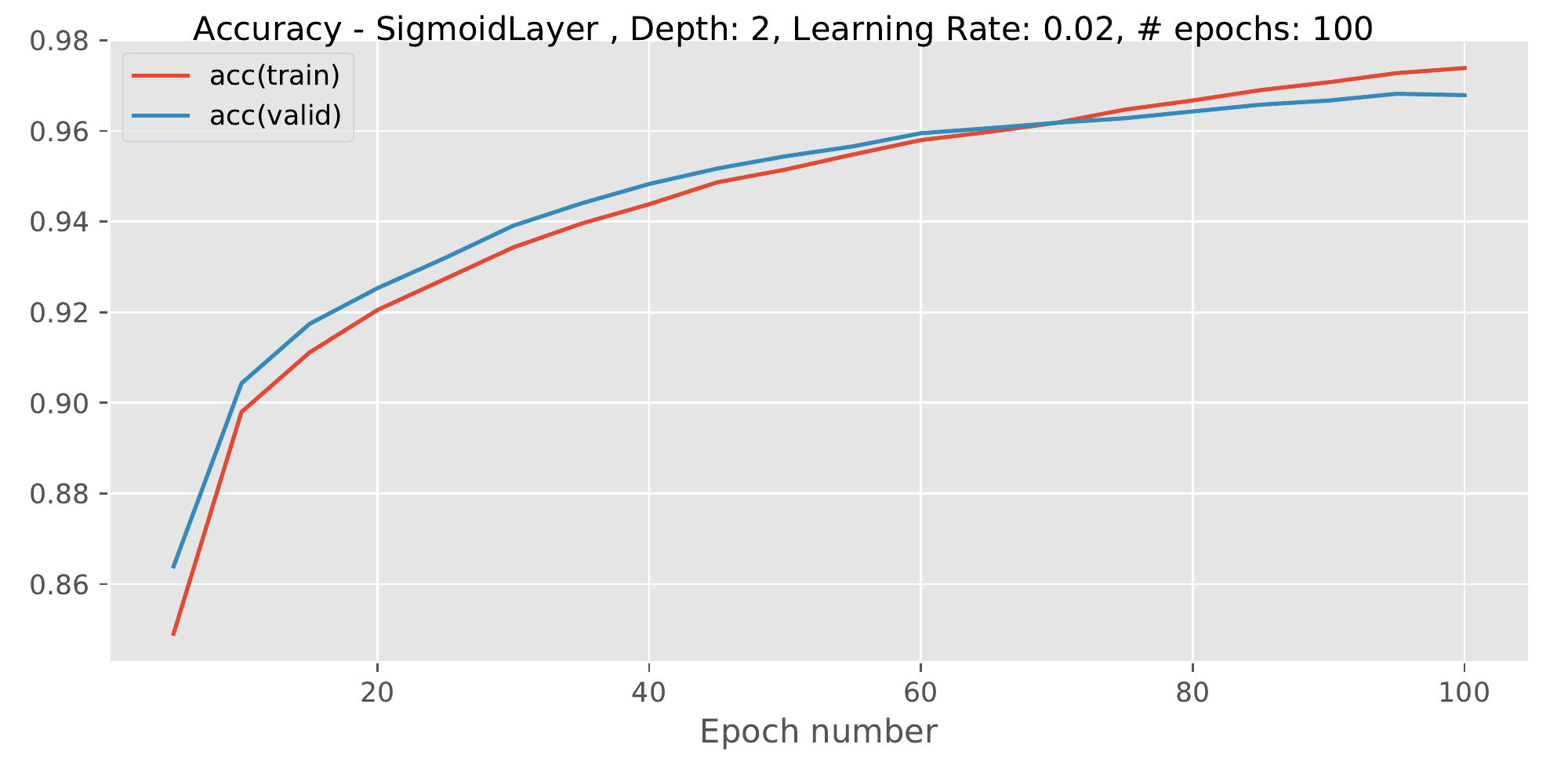}
}
\vskip -3mm
\caption{Sigmoid - Error and accuracy with learning rate 0.02}
\label{fig:SigmoidLayer_depth2_learningrate002_epochs100}
\end{center}
\end{figure}

\begin{figure}[H]
\vskip -3mm
\begin{center}
\subfigure{
\includegraphics[width=38mm, height=30mm]{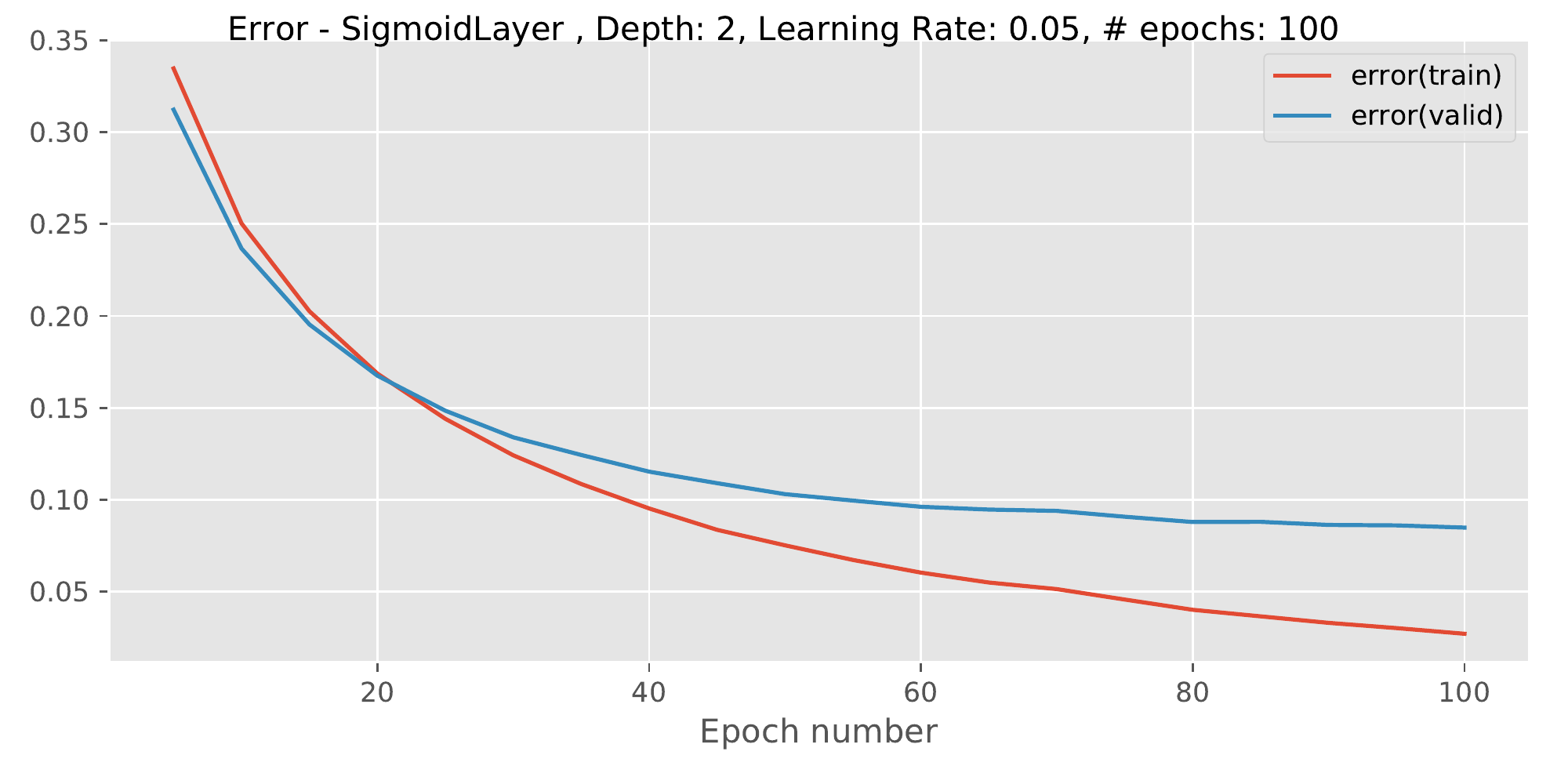}}
\subfigure{\includegraphics[width=38mm, height=30mm]{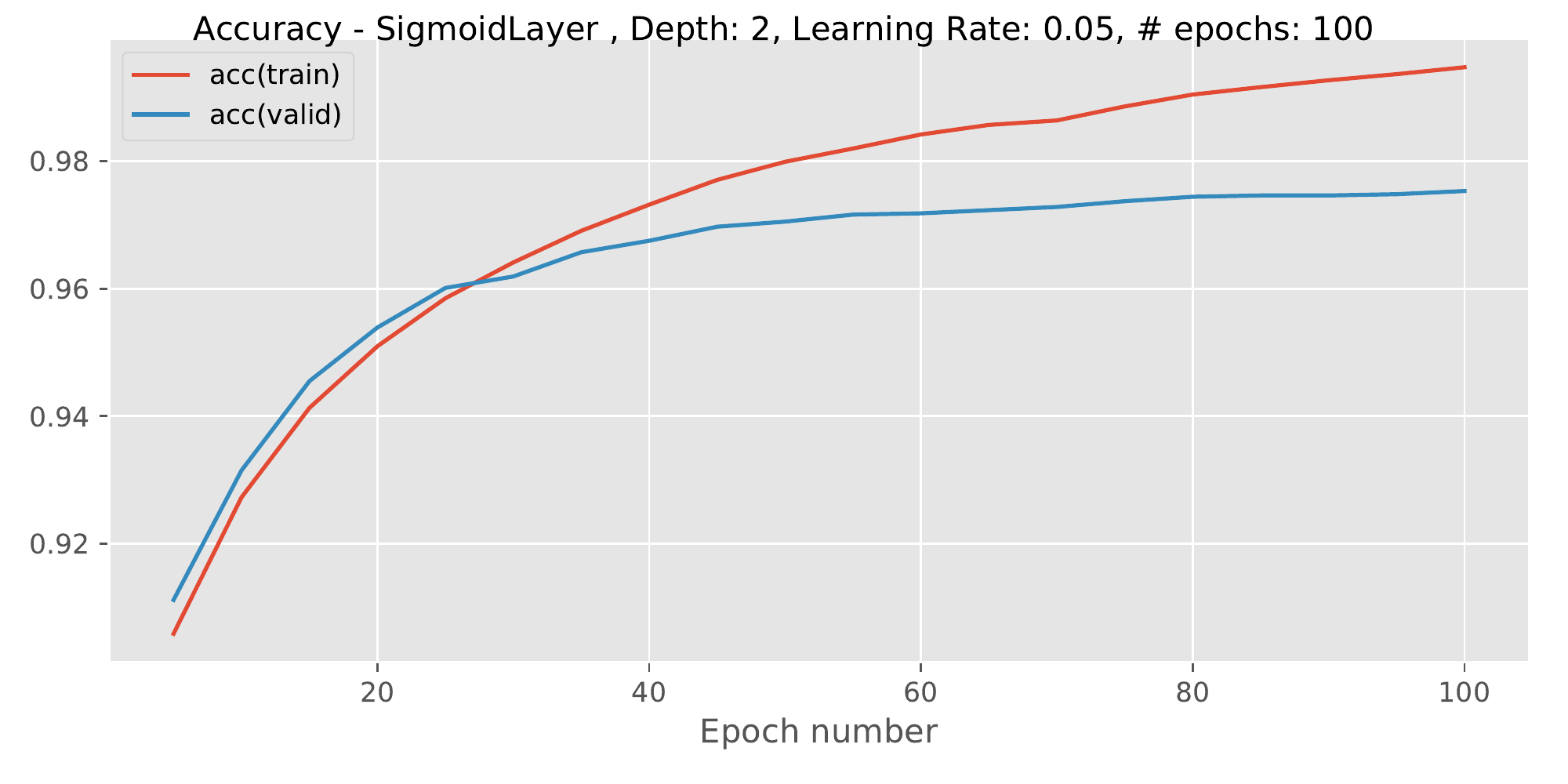}
}
\vskip -3mm
\caption{Sigmoid - Error and accuracy with learning rate 0.05}
\label{fig:SigmoidLayer_depth2_learningrate005_epochs100}
\end{center}
\end{figure}

\begin{figure}[H]
\vskip -3mm
\begin{center}
\subfigure{
\includegraphics[width=38mm, height=30mm]{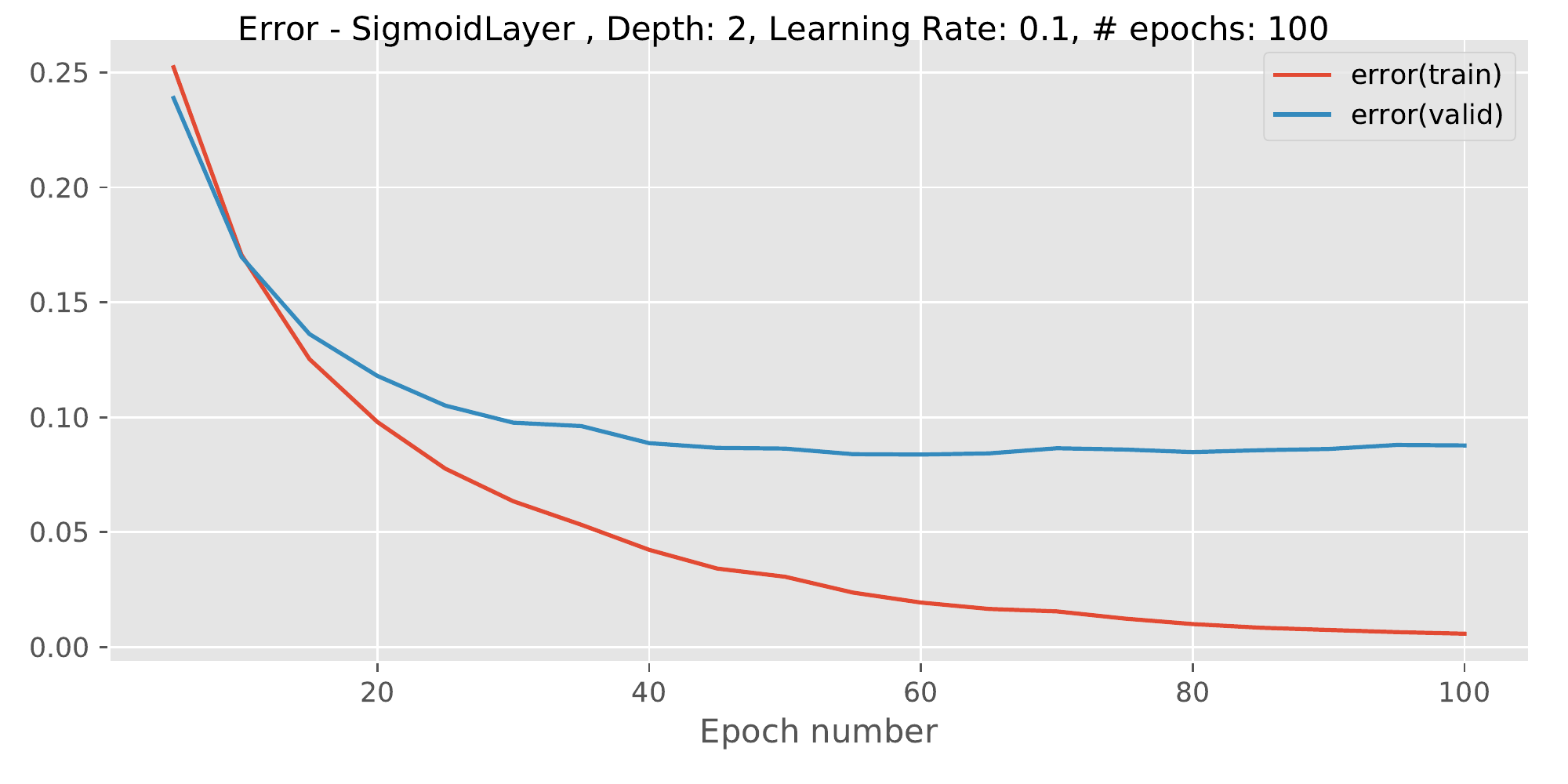}}
\subfigure{\includegraphics[width=38mm, height=30mm]{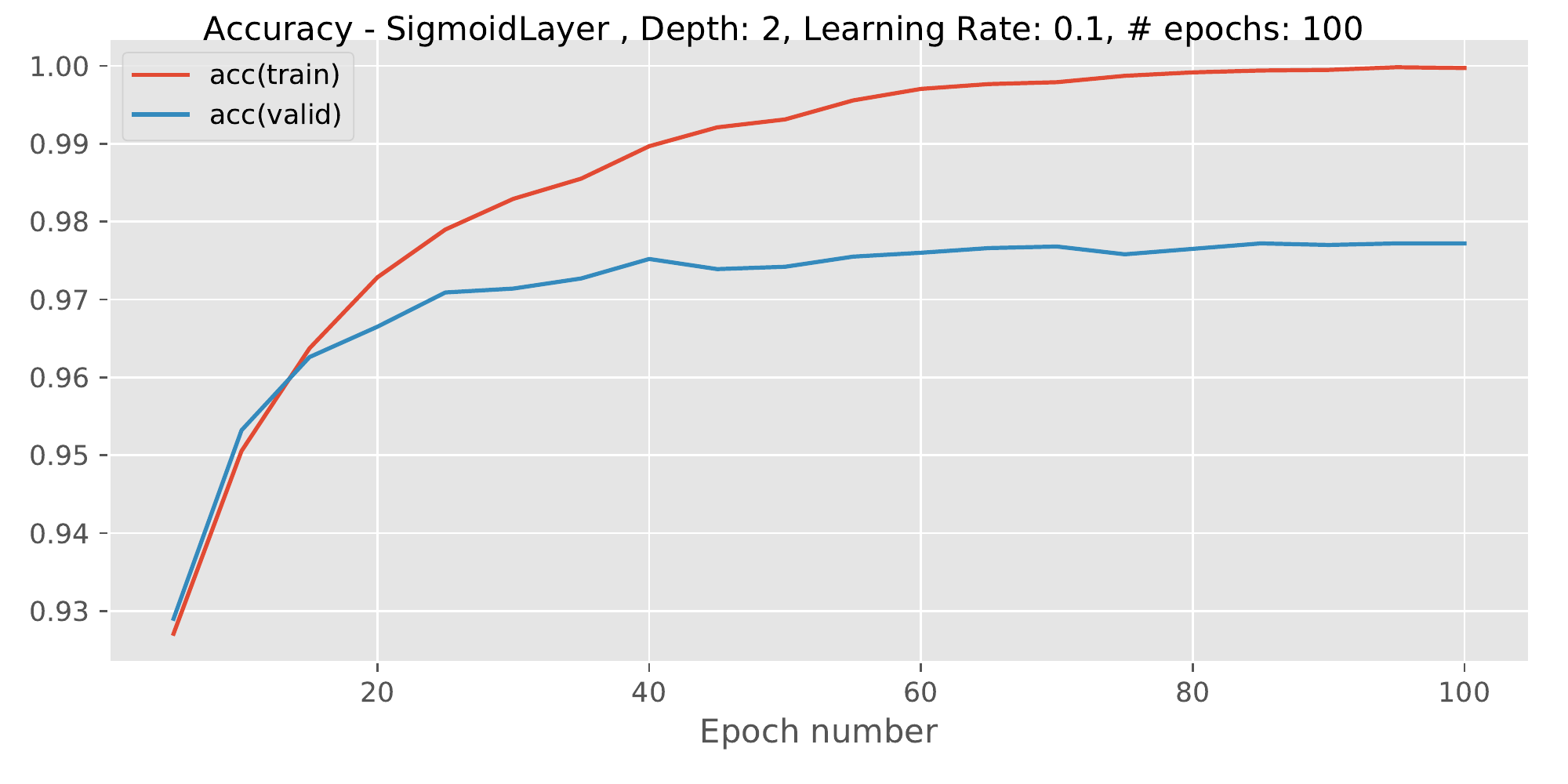}
}
\vskip -3mm
\caption{Sigmoid - Error and accuracy with learning rate 0.10}
\label{fig:SigmoidLayer_depth2_learningrate01_epochs100}
\end{center}
\end{figure}

Reported below in table~\ref{tab:sigmoid-table} are the values achieved for the Sigmoid activation function for the three learning rates.
Figure~\ref{fig:SigmoidLayer_depth2_learningrate002_epochs100}, figure~\ref{fig:SigmoidLayer_depth2_learningrate005_epochs100}, and figure~\ref{fig:SigmoidLayer_depth2_learningrate01_epochs100} show, in order of learning rates, the different values achieved for error (left) and accuracy (right) across all 100 epochs.

We notice that, as the training error decreases, the validation error always decreases, suggesting that the model does not contain unnecessary complexity that lead to overfitting.

We also notice that, as our learning rate increases, both our training error and our accuracy increase. This can be justified by the following two reasons:
\begin{itemize}
  \item With smaller learning rates, gradient descent is more likely to get stuck in local minima.
  \item Smaller learning rates usually require more epochs before an optimal solution is found. With all three learning rates, however we use the same  number of epochs (100). 
\end{itemize}

\begin{table}[H]
\begin{center}
\begin{small}
\begin{sc}
\begin{tabular}{lcccr}
\hline
\abovespace\belowspace
LR & Train error & Valid error & Train acc & Final acc \\
\hline
\abovespace
0.020 &  9.36e-02 &  1.14e-01 &  0.97 &  0.968 \\
0.050 &  2.70e-02 &  8.48e-02 &  0.99 &  0.975 \\
0.100 &  5.77e-03 &  8.77e-02 &  1.00 &  0.977 \\

\hline
\end{tabular}
\end{sc}
\end{small}
\caption{Sigmoid layers: Errors and accuracies for different learning rates}
\label{tab:sigmoid-table}
\end{center}
\end{table}

\begin{figure}[H]
\vskip -3mm
\begin{center}
\centerline{\includegraphics[width=\columnwidth]{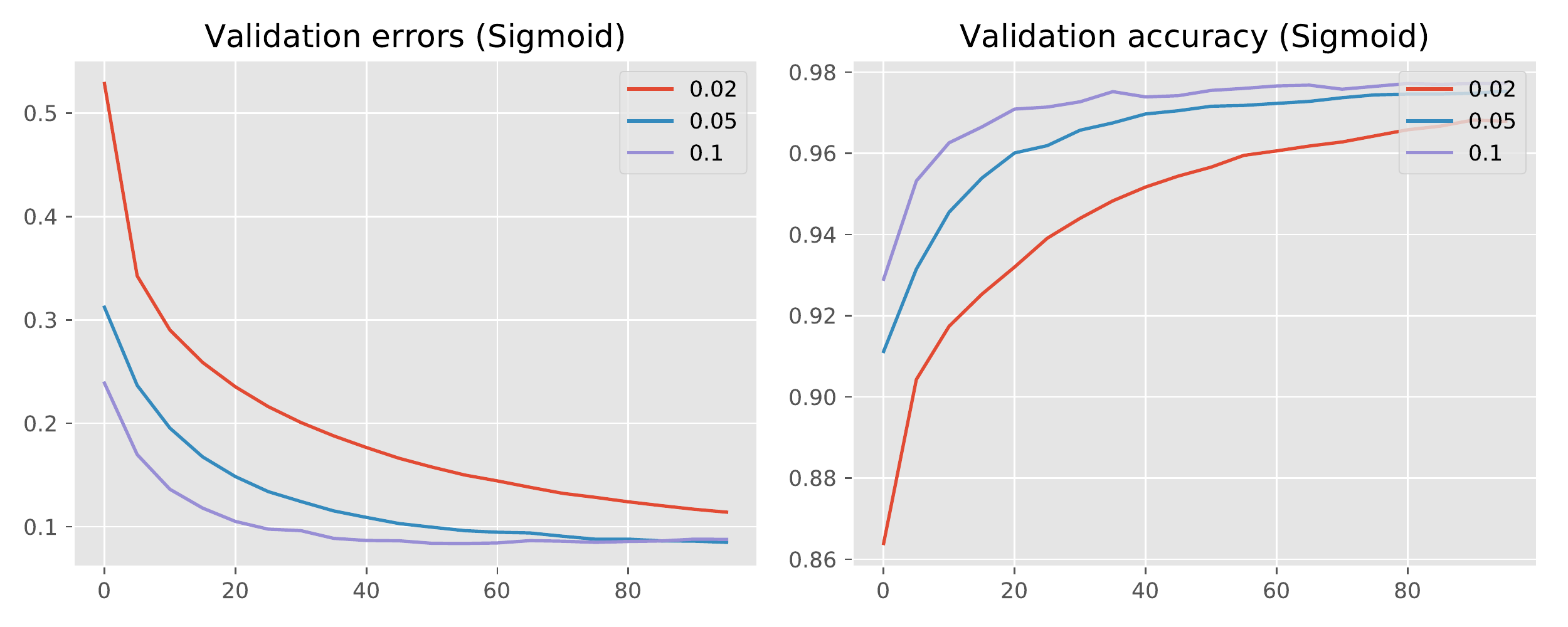}}
\caption{Sigmoid - Validation error and accuracy of different learning rates}
\label{fig:comparison_sigmoid}
\end{center}
\vskip -3mm
\end{figure}

In figure ~\ref{fig:comparison_sigmoid} we visualise the performance, with regards to validation error and validation accuracy of our models with three different learning rates.

\subsection{ReLu}
Rectified Linear Units, our second activation function, is our other baseline model. From table~\ref{tab:relu-table} we see that it already seems to achieve better results than our sigmoid model.
The sigmoid activation function model likely incurred incurred into the vanishing gradient problem \citep{DBLP:journals/corr/abs-1211-5063}, where a big amount of possible inputs precisely \(2^{(28*28)}\) are 'squashed' into a relatively small range: (0, 1), so the computed gradients are small as a result.
\begin{table}[H]
\begin{center}
\begin{small}
\begin{sc}
\begin{tabular}{lcccr}
\hline
\abovespace\belowspace
LR & Train error & Valid error & Train acc & Final acc \\
\hline
\abovespace
0.020 &  1.99e-03 &  1.04e-01 &  1.00 &  0.977 \\
0.050 &  3.99e-04 &  1.12e-01 &  1.00 &  0.978 \\
0.100 &  1.46e-04 &  1.18e-01 &  1.00 &  0.979 \\
\hline
\end{tabular}
\end{sc}
\end{small}
\caption{ReLu layers: Errors and accuracies for different learning rates}
\label{tab:relu-table}
\end{center}
\end{table}

\begin{figure}[H]
\begin{center}
\subfigure{
\includegraphics[width=38mm, height=30mm]{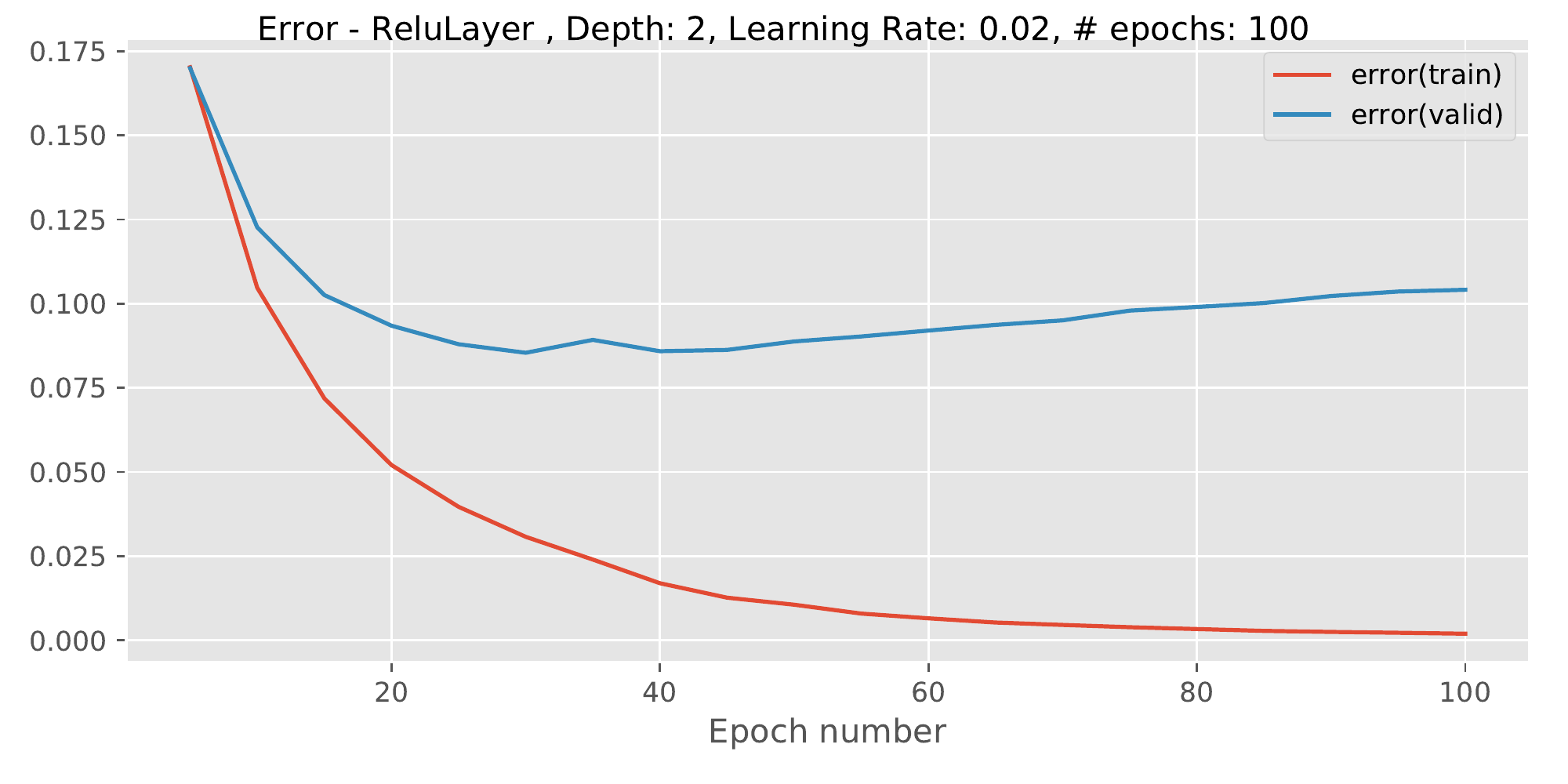}}
\subfigure{\includegraphics[width=38mm, height=30mm]{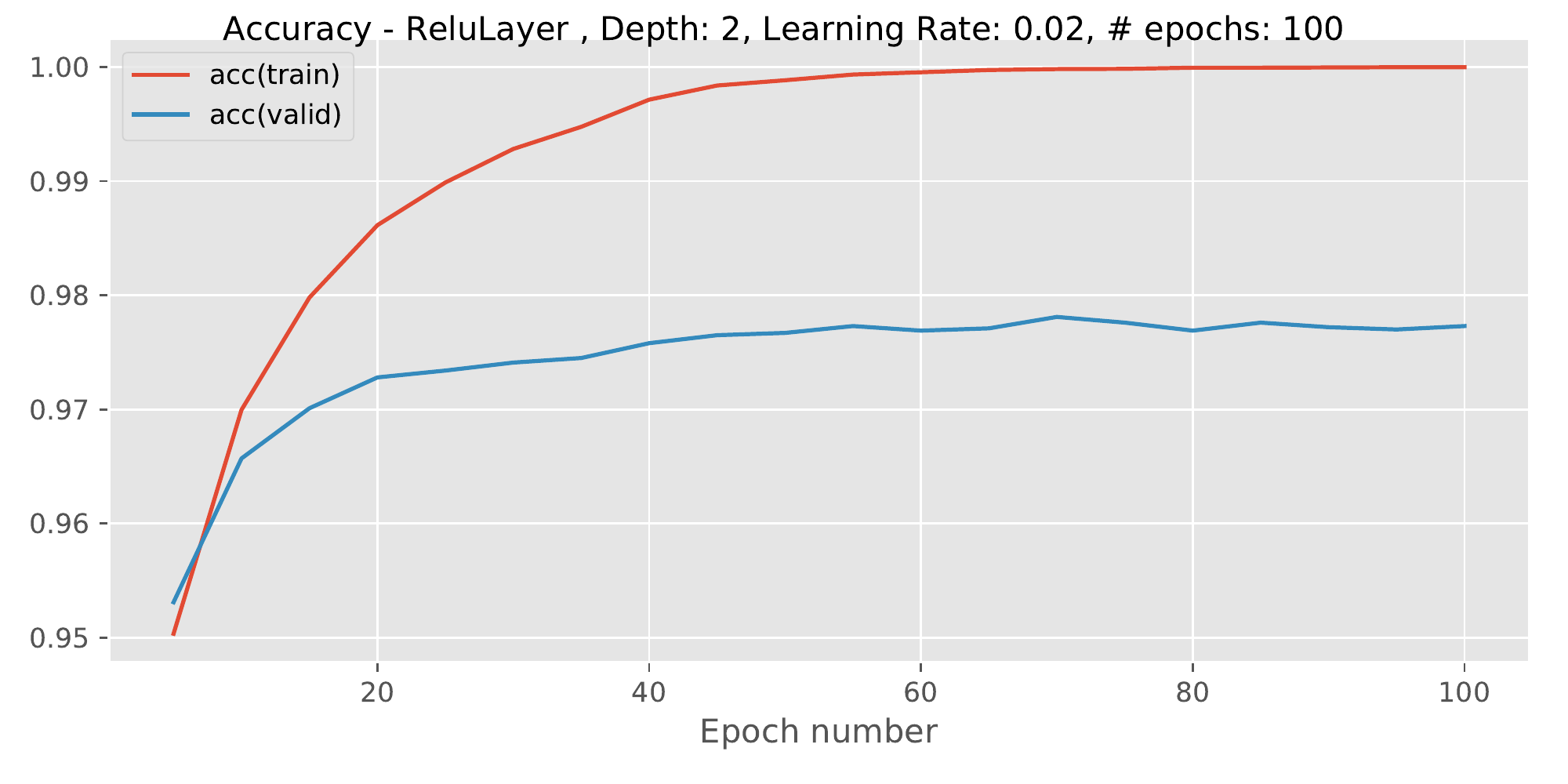}
}
\vskip -3mm
\caption{ReLu - Error and accuracy with learning rate 0.02}
\label{fig:ReluLayer_depth2_learningrate002_epochs100}
\end{center}
\end{figure}

\begin{figure}[H]
\vskip -3mm
\begin{center}
\subfigure{
\includegraphics[width=38mm, height=30mm]{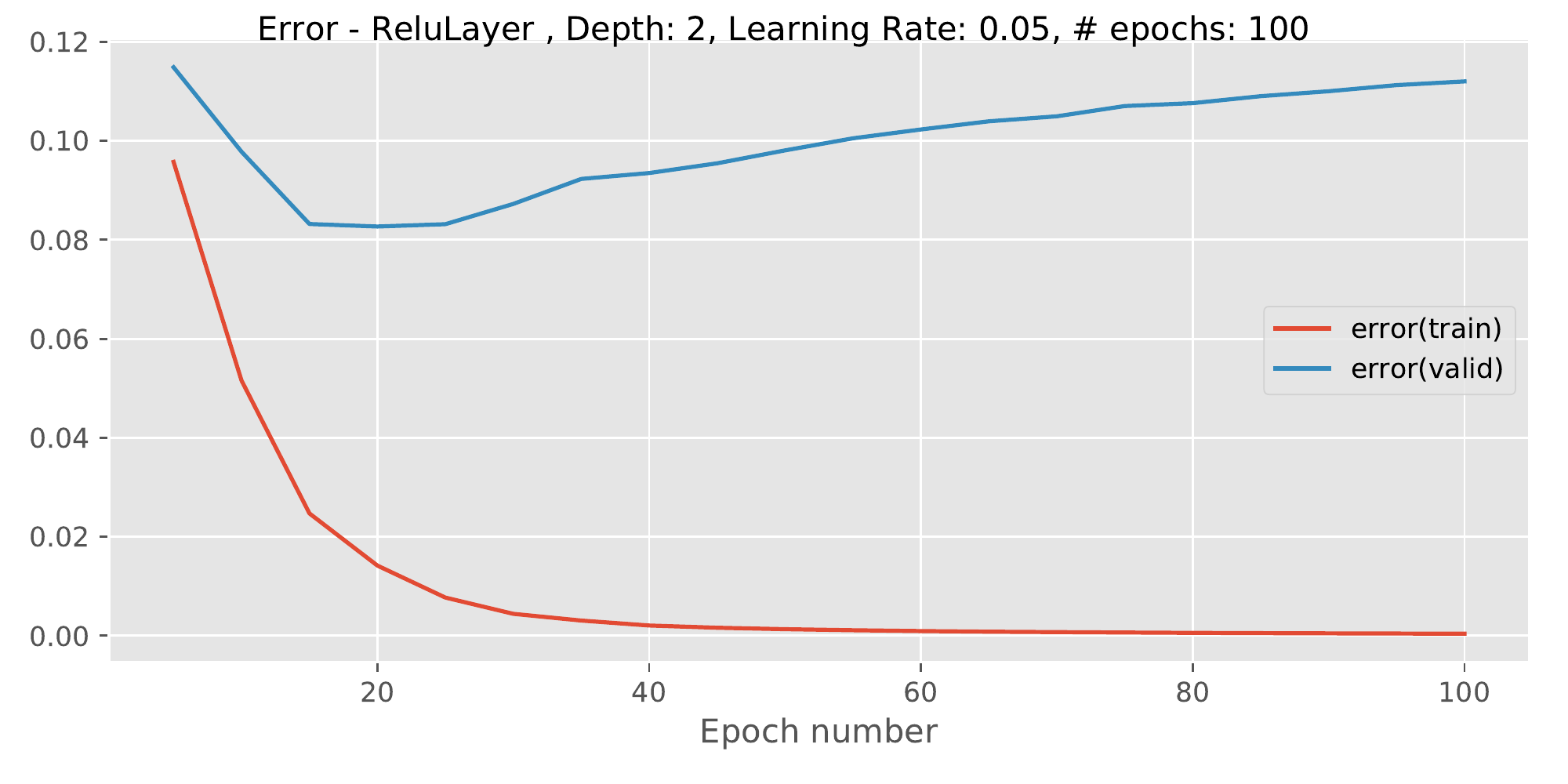}}
\subfigure{\includegraphics[width=38mm, height=30mm]{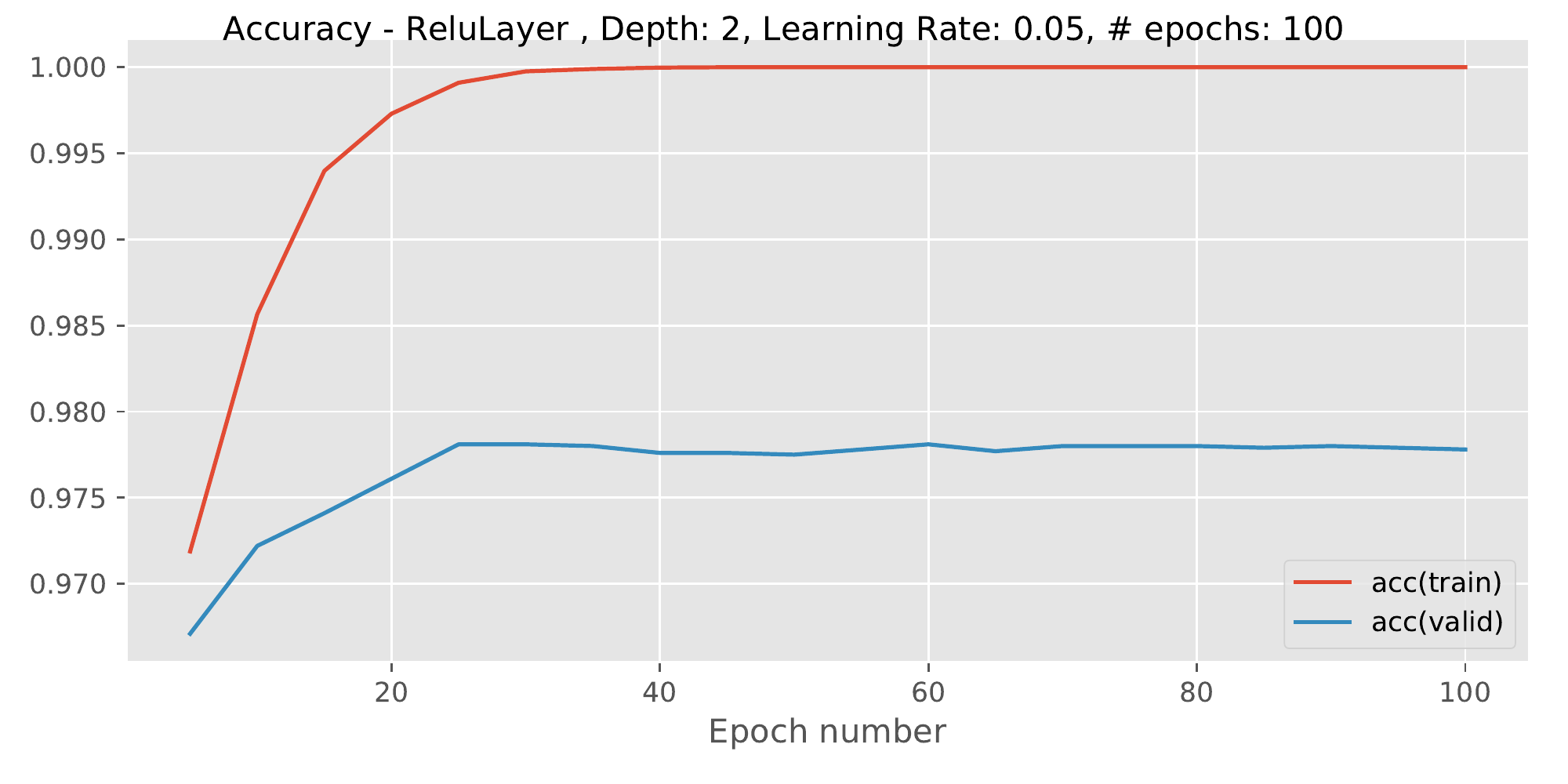}
}
\vskip -3mm
\caption{ReLu - Error and accuracy with learning rate 0.05}
\label{fig:ReluLayer_depth2_learningrate005_epochs100}
\end{center}
\end{figure}

We also notice a stark contrast with the Sigmoid model during training. Our ReLu models tend to find an optimal solution much before our 100th epoch (around 15-25th epoch, more precisely), where our validation error starts increasing, while the training error decreases. This is a sign of overfitting, and could have been address with an early stop at our 15-25th epoch.
Figure~\ref{fig:ReluLayer_depth2_learningrate002_epochs100}, figure~\ref{fig:ReluLayer_depth2_learningrate005_epochs100}, and figure~\ref{fig:ReluLayer_depth2_learningrate005_epochs100} show this behaviour.

\begin{figure}[H]
\vskip -3mm
\begin{center}
\subfigure{
\includegraphics[width=38mm, height=30mm]{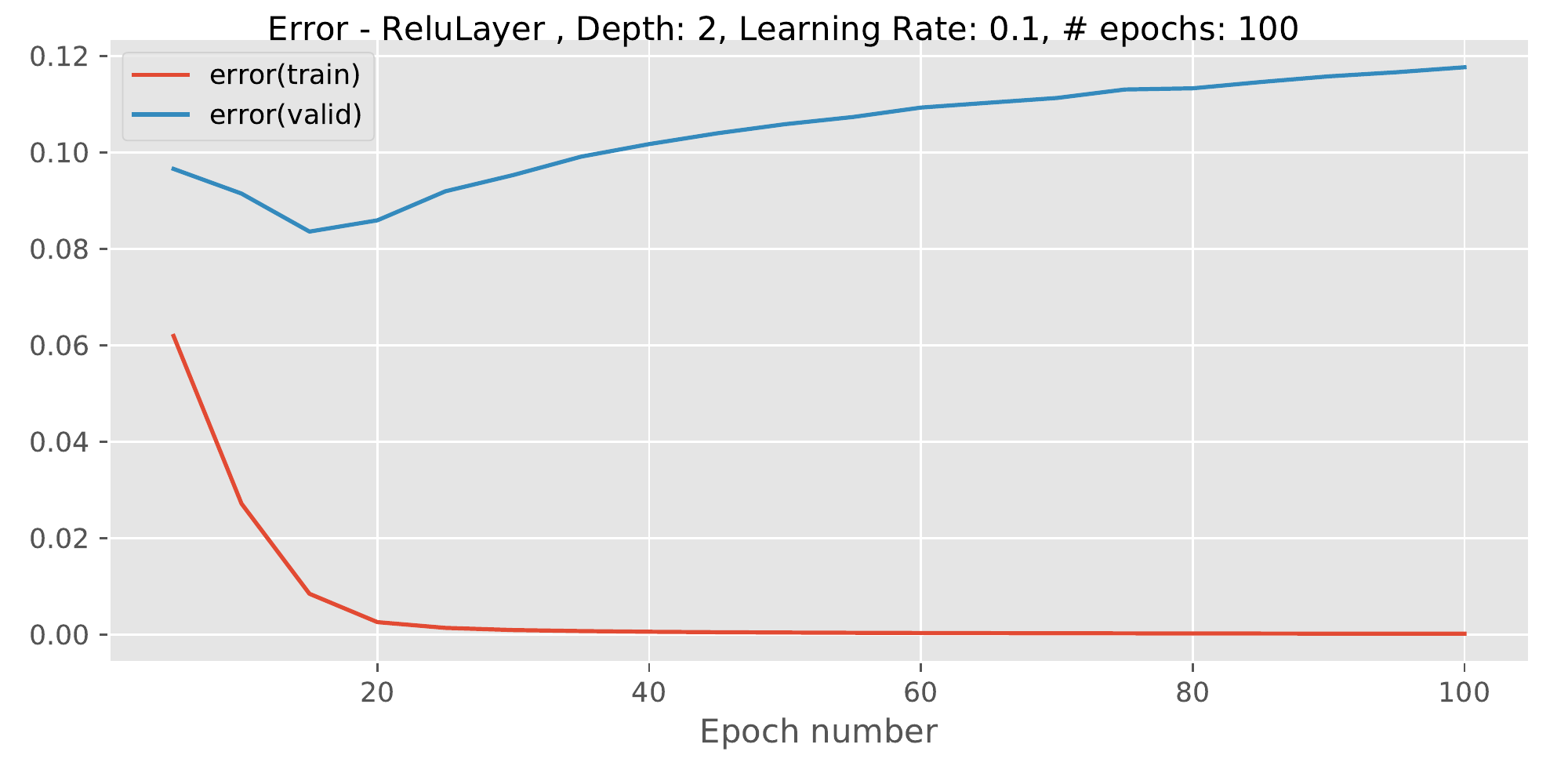}}
\subfigure{\includegraphics[width=38mm, height=30mm]{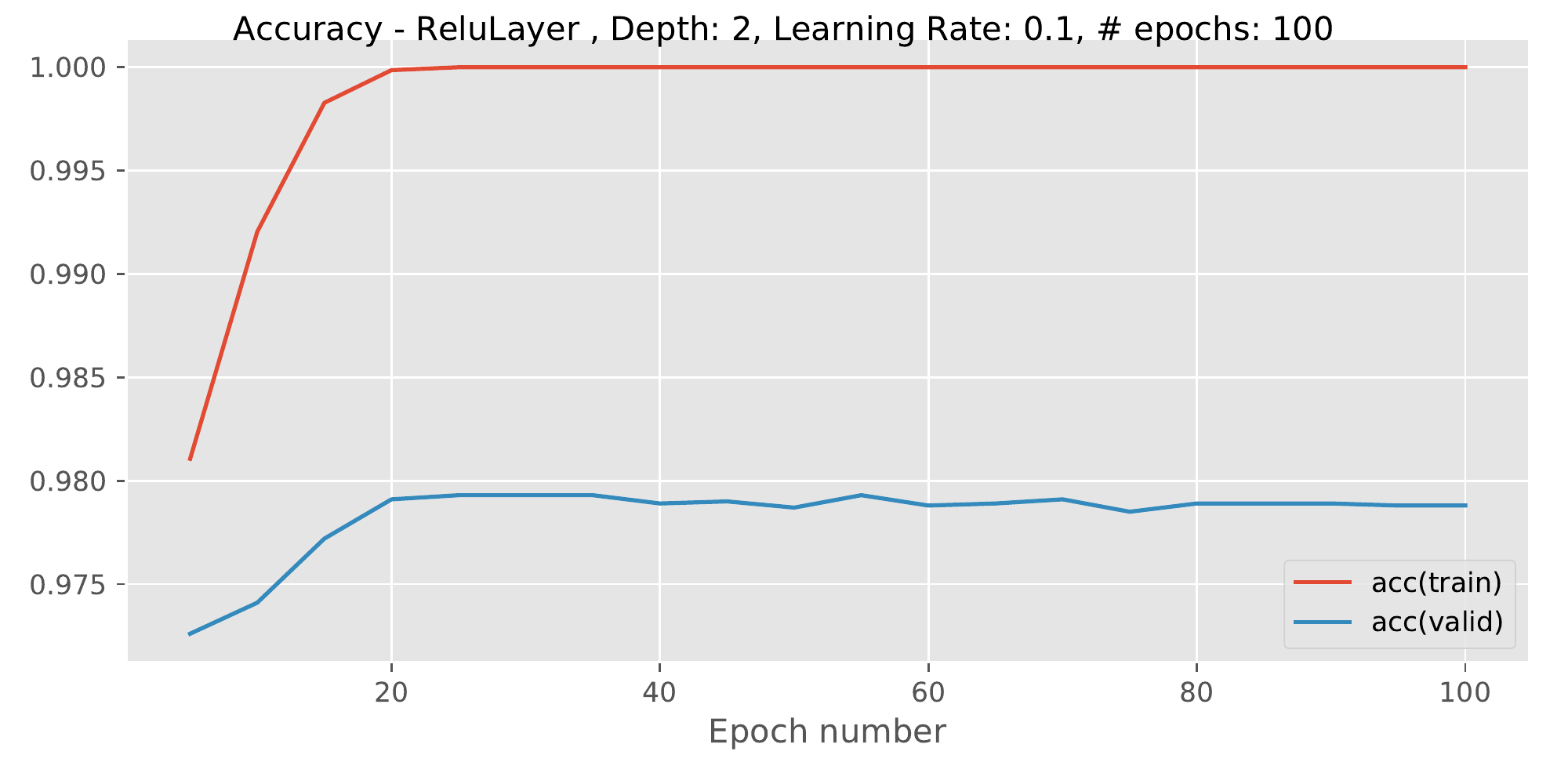}
}
\vskip -3mm
\caption{ReLu - Error and accuracy with learning rate 0.10}
\label{fig:ReluLayer_depth2_learningrate01_epochs100}
\end{center}
\end{figure}

\begin{figure}[H]
\vskip -3mm
\begin{center}
\centerline{\includegraphics[width=\columnwidth]{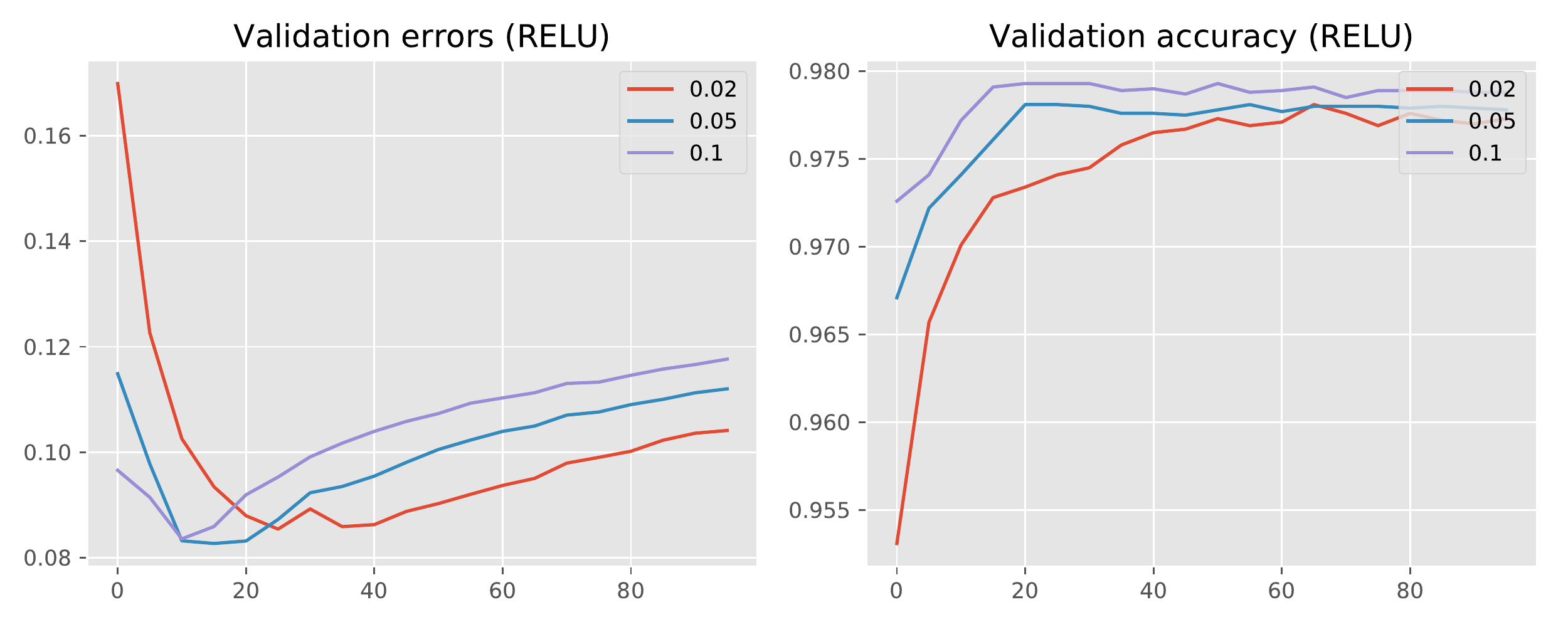}}
\caption{ReLu - Validation error and accuracy of different learning rates}
\label{fig:comparison_relu}
\end{center}
\vskip -3mm
\end{figure} 

\subsection{Other activation functions} \label{otherfn}
In this subsection we report the experimental results with our other three activation functions.
As with our baseline models, we train three models for each activation function, corresponding to the same three learning rates.
For the purpose of not cluttering the section with graphs, we add the detailed performance of each model's learning rate to section \ref{appendixA} (Appendix A). 

Here, we will only report the statistics at the 100th epoch and a summary graph of the performance for each learning rate.

All three model seem to perform equivalently to the ReLu model, and better than the Sigmoid model. The final accuracies at the 100th epoch range from 0.977 to 0.980. The best accuracy is achieved by the Leaky ReLu model, with a final validation accuracy of 0.980, corresponding to the lowest training error of 0.000147.

It is important to note that all three models actually achieved the lowest validation error at a much earlier epoch, just like the ReLu model did.
All three models, in fact, show signs of overfitting (validation error increases, while training error goes to 0). We can infer that, potentially, the validation accuracies could be much greater and adapt better to new data if early stopping was applied.

The final results are in the same range because all three models alleviate the vanishing gradient problem which the Sigmoid activation function tends to incur into. In fact, Leaky ReLu and SELU and ELU all have negative values that help achieve mean shifts toward zero. This justifies the

\textbf{Leaky ReLu:}
\begin{table}[H]
\begin{center}
\begin{small}
\begin{sc}
\begin{tabular}{lcccr}
\hline
\abovespace\belowspace
LR & Train error & Valid error & Train acc & Final acc \\
\hline
\abovespace
0.020 &  2.02e-03 &  1.03e-01 &  1.00 &  0.977 \\
0.050 &  4.07e-04 &  1.13e-01 &  1.00 &  0.978 \\
0.100 &  1.47e-04 &  1.13e-01 &  1.00 &  0.980 \\
\hline
\end{tabular}
\end{sc}
\end{small}
\caption{Leaky ReLu layers: Errors and accuracies for different learning rates}
\label{tab:lrelu-table}
\end{center}
\end{table}

\begin{figure}[H]
\vskip -3mm
\begin{center}
\centerline{\includegraphics[width=\columnwidth]{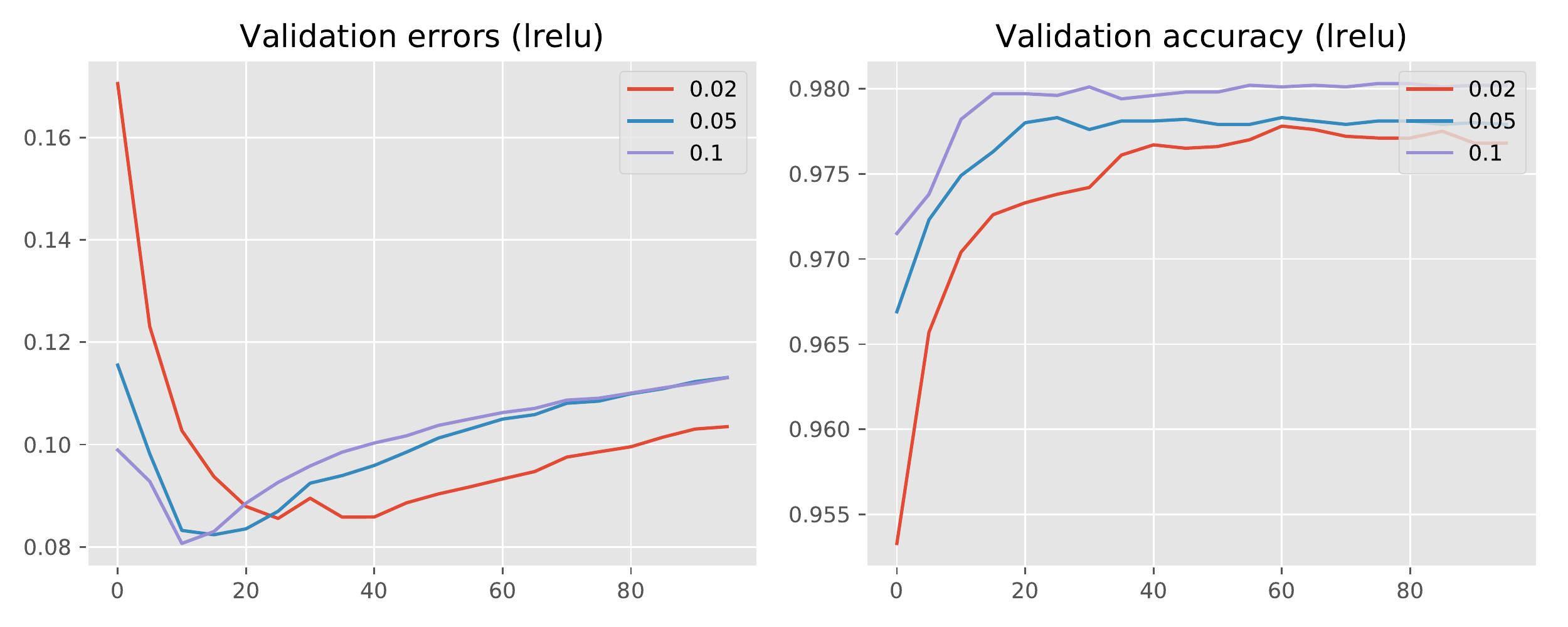}}
\caption{Leaky ReLu - Validation error and accuracy of different learning rates}
\label{fig:comparison_lrelu}
\end{center}
\vskip -3mm
\end{figure} 

\textbf{ELU:}
\begin{table}[H]
\begin{center}
\begin{small}
\begin{sc}
\begin{tabular}{lcccr}
\hline
\abovespace\belowspace
LR & Train error & Valid error & Train acc & Final acc \\
\hline
\abovespace
0.020 &  3.62e-03 &  1.03e-01 &  1.00 &  0.977 \\
0.050 &  5.84e-04 &  1.18e-01 &  1.00 &  0.978 \\
0.100 &  1.98e-04 &  1.22e-01 &  1.00 &  0.979 \\
\hline
\end{tabular}
\end{sc}
\end{small}
\caption{ELU layers: Errors and accuracies for different learning rates}
\label{tab:elu-table}
\end{center}
\vskip -3mm
\end{table}

\begin{figure}[H]
\vskip -3mm
\begin{center}
\centerline{\includegraphics[width=\columnwidth]{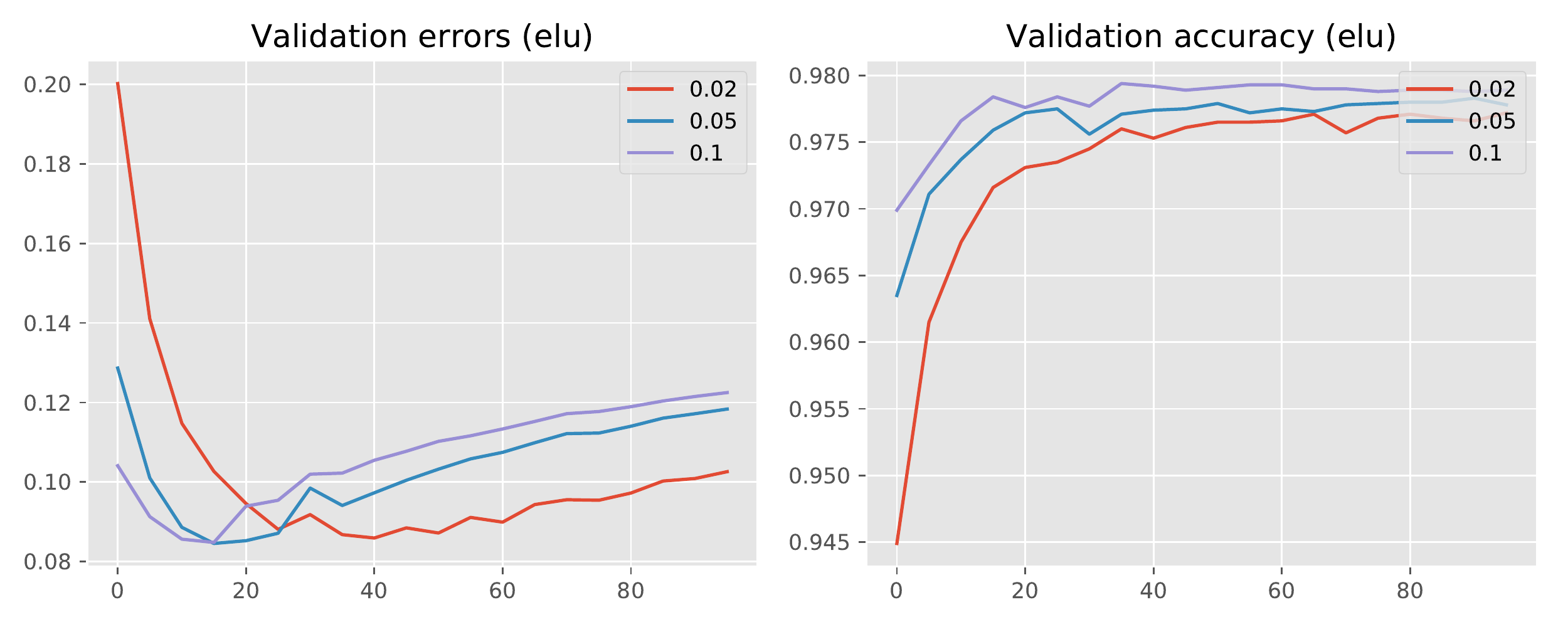}}
\caption{ELU - Validation error and accuracy of different learning rates}
\label{fig:comparison_elu}
\end{center}
\vskip -3mm
\end{figure} 

\textbf{SELU:}
\begin{table}[H]
\vskip 3mm
\begin{center}
\begin{small}
\begin{sc}
\begin{tabular}{lcccr}
\hline
\abovespace\belowspace
LR & Train error & Valid error & Train acc & Final acc \\
\hline
\abovespace
0.020 &  2.53e-03 &  1.09e-01 &  1.00 &  0.976 \\
0.050 &  5.28e-04 &  1.19e-01 &  1.00 &  0.977 \\
0.100 &  1.95e-04 &  1.26e-01 &  1.00 &  0.978 \\\hline
\end{tabular}
\end{sc}
\end{small}
\caption{SELU layers: Errors and accuracies for different learning rates}
\label{tab:selu-table}
\end{center}
\vskip -3mm
\end{table}

\begin{figure}[H]
\vskip -3mm
\begin{center}
\centerline{\includegraphics[width=\columnwidth]{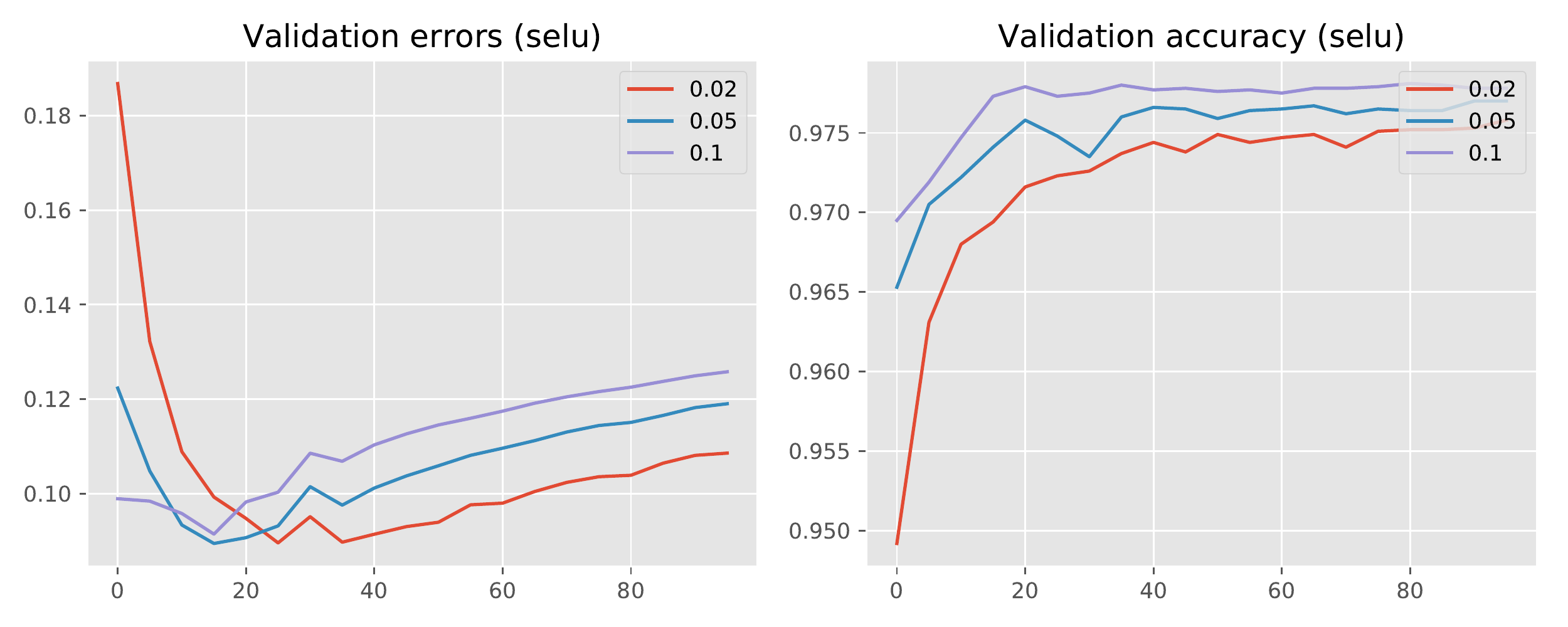}}
\caption{SELU - Validation error and accuracy of different learning rates}
\label{fig:comparison_selu}
\end{center}
\vskip -3mm
\end{figure} 

\section{Deep neural network experiments}
\label{sec:dnnexpts}
This section reports on experiments on deeper networks for MNIST. Specifically, the first set of experiments explores the impact of the depth of the network, by varying the number of hidden layers from 2 to 8. A second set of experiments is aims at  comparing different approaches to weight initialisation.

Experiments in these sections have been carried with a fixed model chosen from the first part of the report. We use a model with 4 hidden ELU layers, and a learning rate of 0.02. 

This selection is not arbitrary. We selected ELU over other activation functions not only because it achieve slightly achieves better experimental results in the previous set of experiments, but also because we identified a faster training time for these layers, which came in useful when training deeper models. The faster training time is justified in \citep{DBLP:journals/corr/ClevertUH15}.

Additionally, we chose a learning rate of 0.02 over 0.05 and 0.1 despite it having a slightly lower accuracy. This is to prevent overflows that resulted from our implementation of our softmax layer.

\subsection{Deep models}
\label{sec:deepmodelexp}
Table~\ref{tab:deep-table} reports the training/validation errors and accuracy for the models with ELU hidden layers, varying from a depth of 3 to 8. We use the one with 2 ELU hidden layers from the previous section as a base model.

The first thing that can be noticed is that we can strictly achieve a lower training error as we increase the depth, and thus the complexity of our layer. In fact, we can achieve, in our models with 7-8 hidden layers, a two-fold decrease of the training error, which is almost annulled after the 100th epoch.

This result does not always correspond to a better model--in fact as the training error decreases, the validation error or the accuracy do not improve. The baseline model is the one that achieves the lowest validation error, which is a sign that our data does not meaningfully benefit from the added complexity. Furthermore, we notice that the accuracy improves almost insignificantly from our baseline model, achieving at best a 0.003 increase.

\begin{table}[H]
\vskip 3mm
\begin{center}
\begin{small}
\begin{sc}
\begin{tabular}{lcccr}
\hline
\abovespace\belowspace
Depth & Train error & Valid error & Train acc & Final acc \\
\hline
\abovespace
2 & 3.62e-03 & 1.03e-01 & 1.00 & 0.977 \\
3 & 9.85e-04 & 1.19e-01 & 1.00 & 0.977 \\
4 & 3.93e-04 & 1.29e-01 & 1.00 & 0.979 \\
5 & 2.02e-04 & 1.44e-01 & 1.00 & 0.979 \\
6 & 1.22e-04 & 1.48e-01 & 1.00 & 0.979 \\
7 & 8.63e-05 & 1.52e-01 & 1.00 & 0.980 \\
8 & 6.42e-05 & 1.43e-01 & 1.00 & 0.979 \\
\hline
\end{tabular}
\end{sc}
\end{small}
\caption{Errors and accuracies for different depths of an model with ELU hidden layers after 100 epochs}
\label{tab:deep-table}
\end{center}
\vskip -3mm
\end{table}

\begin{figure}[H]
\vskip -3mm
\begin{center}
\centerline{\includegraphics[width=\columnwidth]{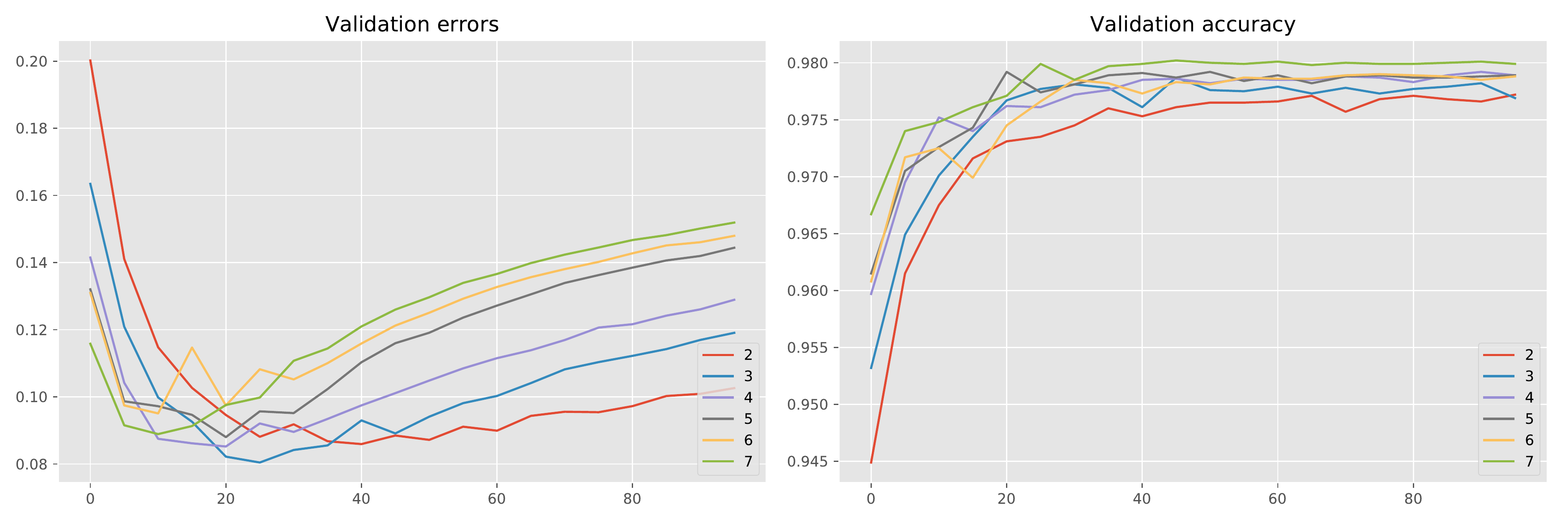}}
\caption{Validation error and accuracy of models with different depths}
\label{fig:comparison_depth}
\end{center}
\vskip -3mm
\end{figure}

Figure~\ref{fig:comparison_depth} shows how the validation error and accuracy vary over the 100 epochs as we change the model's depth. This gives a better picture of how early stopping could have 	actually helped prevent overfitting and come out with a better model.

It is interesting to see how the lowest validation error is achieved by the model with 3 hidden ELU layers at around epoch 25. What happens is that after epoch 25, the validation error of this model increases with a steeper slope than the one with depth 2 (baseline), since we end up overfitting the data more due to the increase complexity. It results, at epoch 100, that its validation error is higher. We can still infer that a slightly more complex, and less overtrained model could achieve better results in terms of validation error.

\subsection{Weight initialisation experiments}
\label{sec:weightinitexp}
In all preceding experiments, our model's weights have been initialised using the Glorot/Xavier initilisation \citep{pmlr-v9-glorot10a}. Following their convention, initial weights are sampled from a uniform distribution whose range depends on the incoming ($n_{in}$) and outcoming ($n_{out}$) connections for units of our hidden layers.

More specifically our weights are sampled in this way:

\begin{equation} \label{faninout}
w_{i} \sim U\Bigg(-\sqrt{\frac{6}{n_{in}+n_{out}}}, \sqrt{\frac{6}{n_{in}+n_{out}}}\Bigg)
\end{equation}

In this section, we report results from using two other initialisation schemes, more specifically weights sampled from equations \ref{fanin} and \ref{fanout}.

\begin{equation} \label{fanin}
  w_{i} \sim U\Bigg(-\sqrt{\frac{3}{n_{in}}}, \sqrt{\frac{3}{n_{in}}}\Bigg)
\end{equation}

\begin{equation} \label{fanout}
  w_{i} \sim U\Bigg(-\sqrt{\frac{3}{n_{out}}}, \sqrt{\frac{3}{n_{out}}}\Bigg)
\end{equation}

In this experiment we once again fix the learning rate, the model's depth and hidden layer types, to have comparable results. In an analogous way as the previous experiments, we use a learning rate of 0.02 and ELU hidden layer types. We also choose to use a deeper layer model (depth 4) to potentially magnify the effect of using different weights.

Table~\ref{tab:fan-table} shows that our baseline model with Glorot Uniform initilisation performs much better than using fan-in and fan-out techniques. Validation errors and accuracies are lower after 100 epochs.

This result was somewhat expected because sampling weights following Glorot/Xavier (equation~\ref{faninout}) is just a combination of equations \ref{fanin} and \ref{fanout}, where both forward propagation and backward propagation are taken into consideration when we calibrate the variance of the input independent of the number of incoming and outcoming connections to a single unit.
 
\begin{table}[H]
\vskip 3mm
\begin{center}
\begin{small}
\begin{sc}
\begin{tabular}{lcccr}
\hline
\abovespace\belowspace
Schema & Train error & Valid error & Train acc & Final acc \\
\hline
\abovespace
Glorot & 3.93e-04 & 1.29e-01 & 1.00 & 0.979 \\
Fan-in & 3.62e-03 & 1.78e-01 & 1.00 & 0.973 \\
Fan-out & 7.76e-04 & 3.46e-01 & 1.00 & 0.955 \\
\hline
\end{tabular}
\end{sc}
\end{small}
\caption{Errors and accuracies for different depths of a model with ELU hidden layers after 100 epochs}
\label{tab:fan-table}
\end{center}
\vskip -3mm
\end{table}

\begin{figure}[H]
\vskip -3mm
\begin{center}
\centerline{\includegraphics[width=\columnwidth]{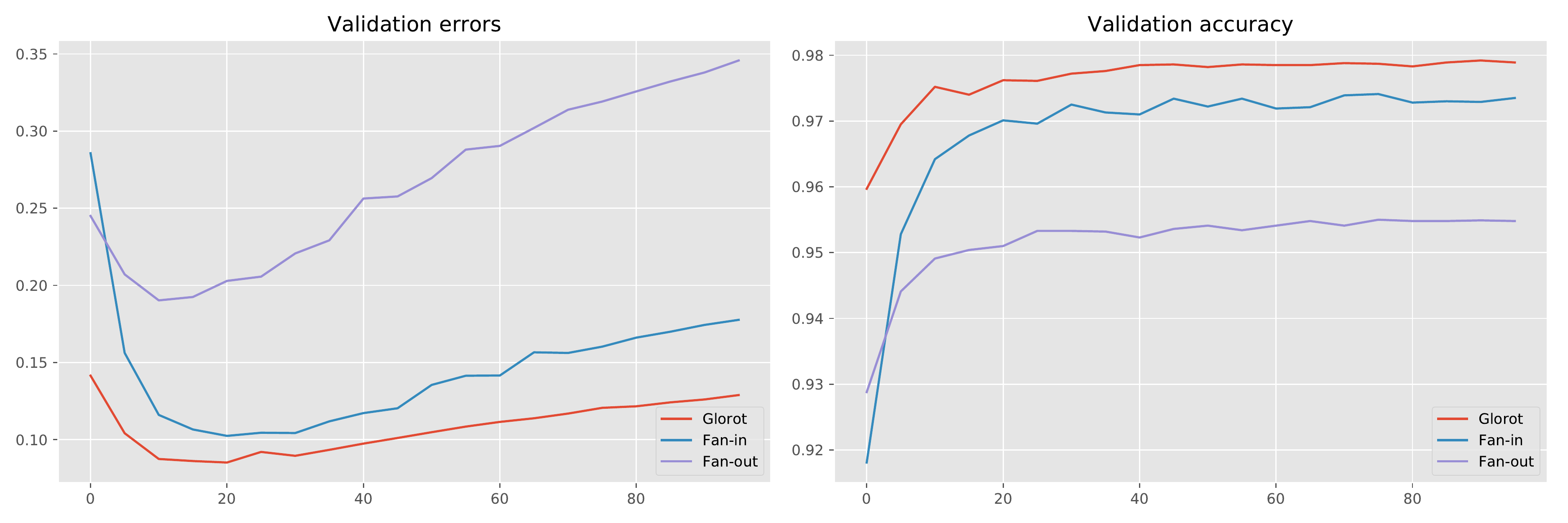}}
\caption{Validation error and accuracy of models with different initialisation schemes}
\label{fig:comparison_fans}
\end{center}
\vskip -3mm
\end{figure}

Figure~\ref{fig:comparison_fans} shows how validation errors and accuracy vary across the 100 epochs. Given the same number of epochs, Glorot initilisation is always superior to both other methods. Also, fan-in initilisation performs better than fan-out, which is prone to overfitting the model in our case.

\section{Conclusions}
\label{sec:concl}
As demonstrated with the first set of experiments, ReLu, Leaky ReLu, ELU and SELU activation functions all yield great results in terms of validation error and accuracy on the MNIST task, with the ELU layer overall performing better than all other models.

Given this results, we found that using an additional hidden layer (for a total of 3 hidden ELU layers) improves the overall performance.
Applying early stopping has proven to yield lower validation error and higher accuracy, as the models, particularly the deeper ones, tend to overfit the training data.

With this model, we also establish that the Glorot/Xavier Uniform initialisation schema yields the best result, compared to techniques that only take into account the incoming or the outcoming connections to a single node in the network.

To conclude, the following model is the one we found to yield the best results:\\
Activation function: ELU \\
Learning rate: 0.02 \\
Number of epochs: 25 \\
Number of hidden layers: 3 \\
Weight initialisation: Glorot Uniform

In further work, we will generalise these results to more complex classification tasks, like CIFAR-10 or CIFAR-100 \citep{krizhevsky2009learning} as well as regression tasks.

\renewcommand\thesubsection{\Alph{subsection}}
\section{Appendix A} \label{appendixA}
In this section we append error and accuracy graphs for different learning rates, as companion data to section \ref{otherfn}.

\textbf{Leaky ReLu:}
\begin{figure}[H]
\vskip -3mm
\begin{center}
\subfigure{
\includegraphics[width=38mm, height=30mm]{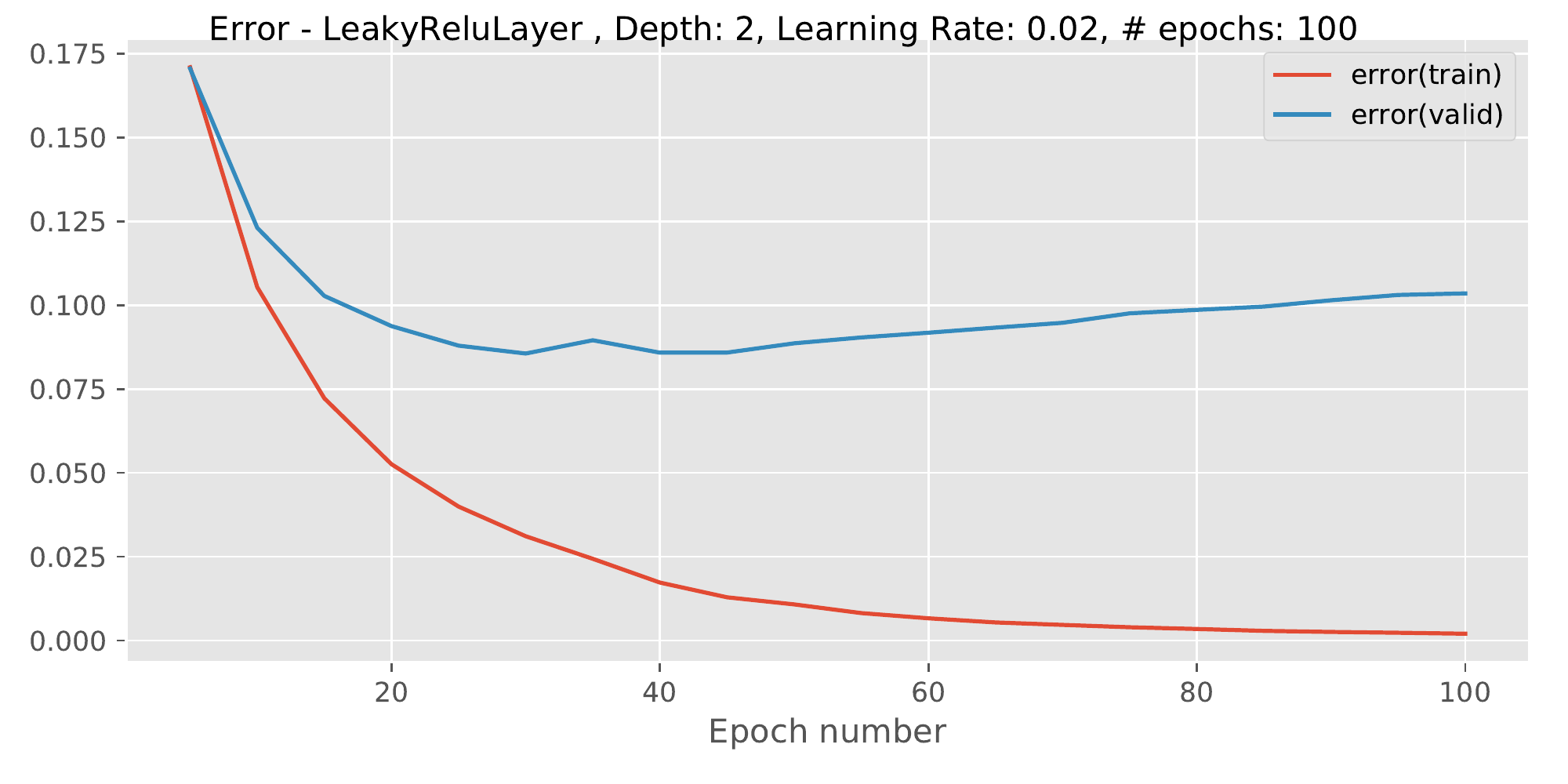}}
\subfigure{\includegraphics[width=38mm, height=30mm]{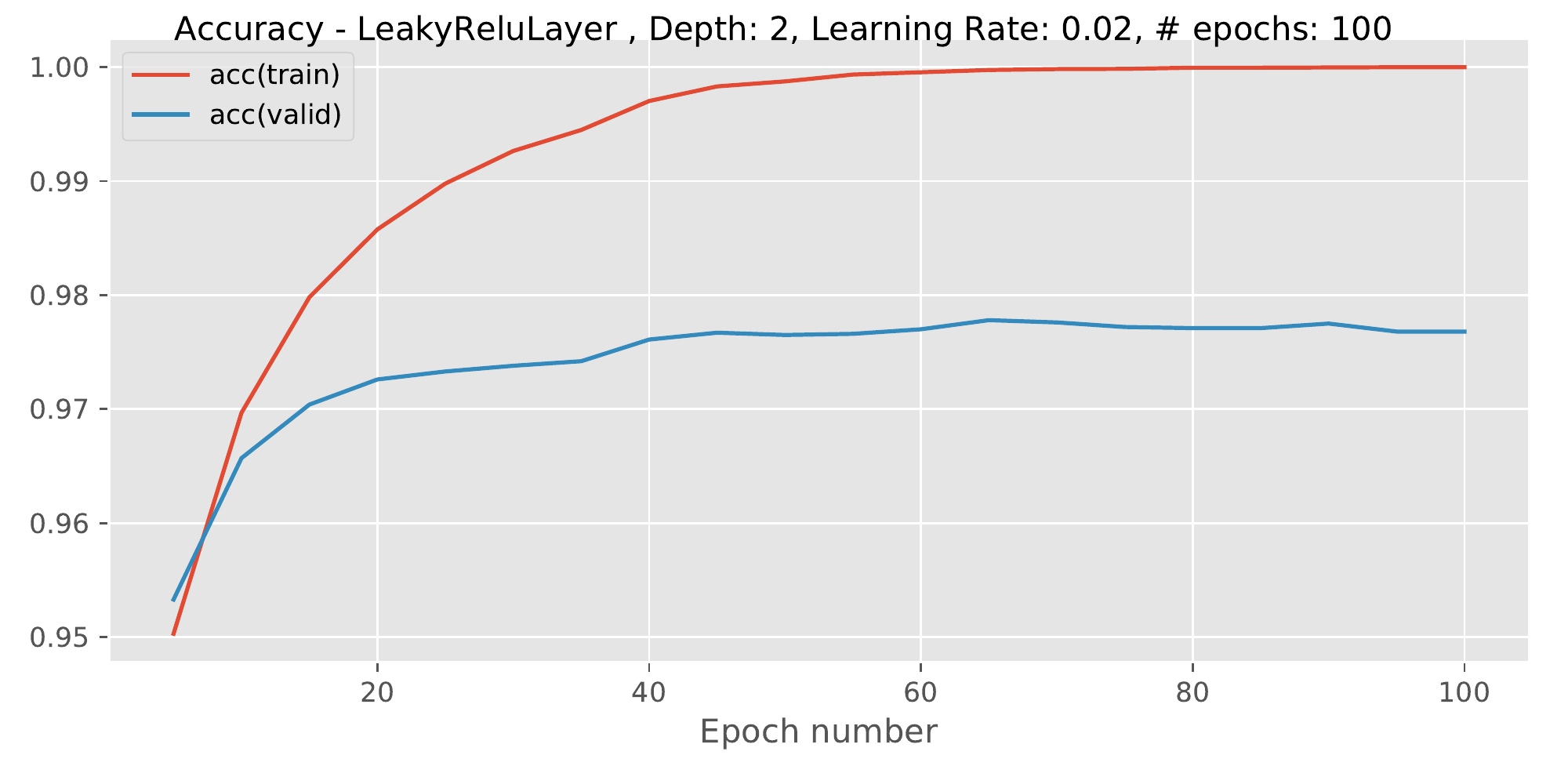}
}
\vskip -3mm
\caption{LeakyRelu - Error and accuracy with learning rate 0.02}
\label{fig:LeakyReluLayer_depth2_learningrate002_epochs100}
\end{center}
\end{figure}

\begin{figure}[H]
\vskip -3mm
\begin{center}
\subfigure{
\includegraphics[width=38mm, height=30mm]{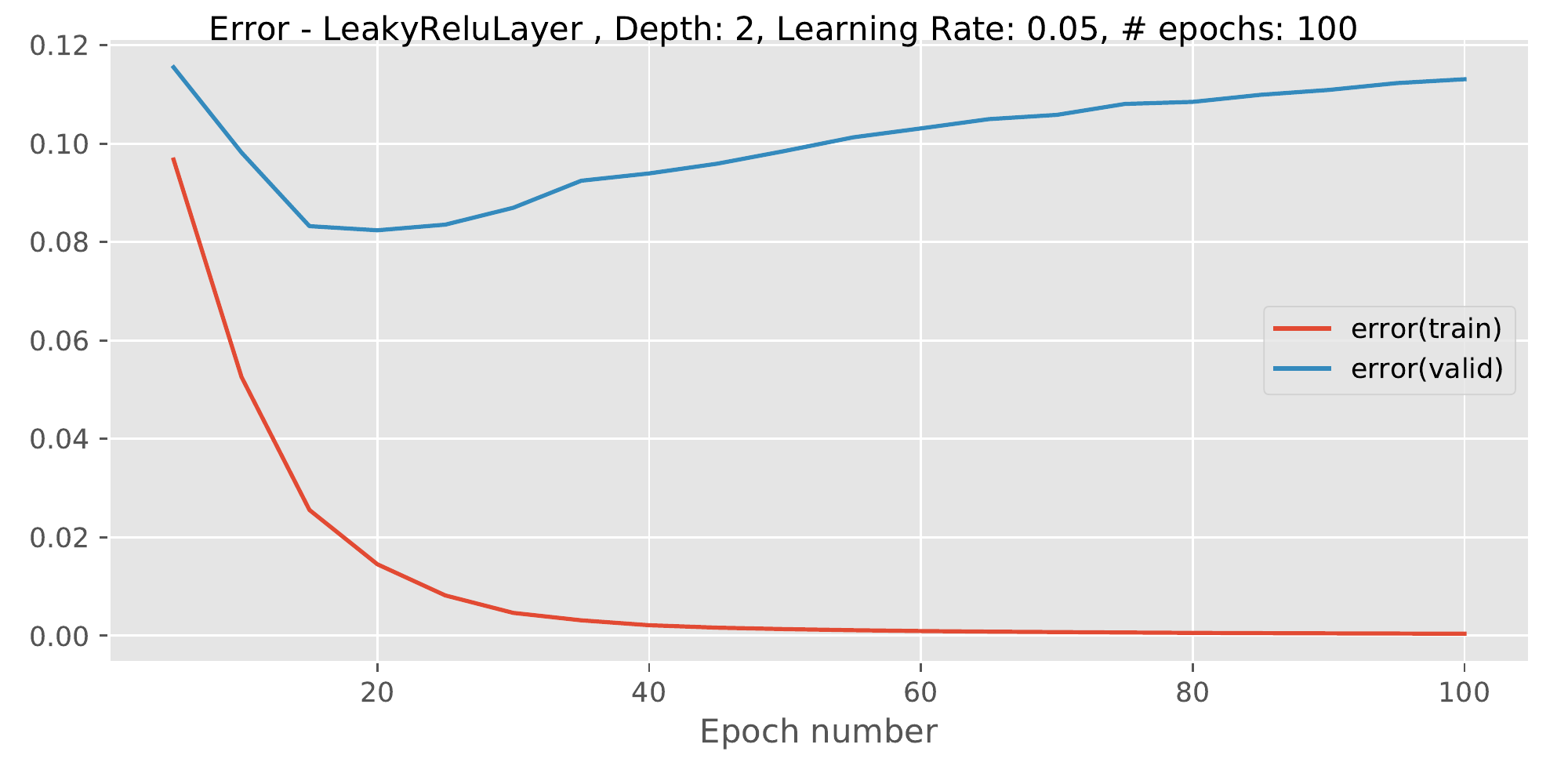}}
\subfigure{\includegraphics[width=38mm, height=30mm]{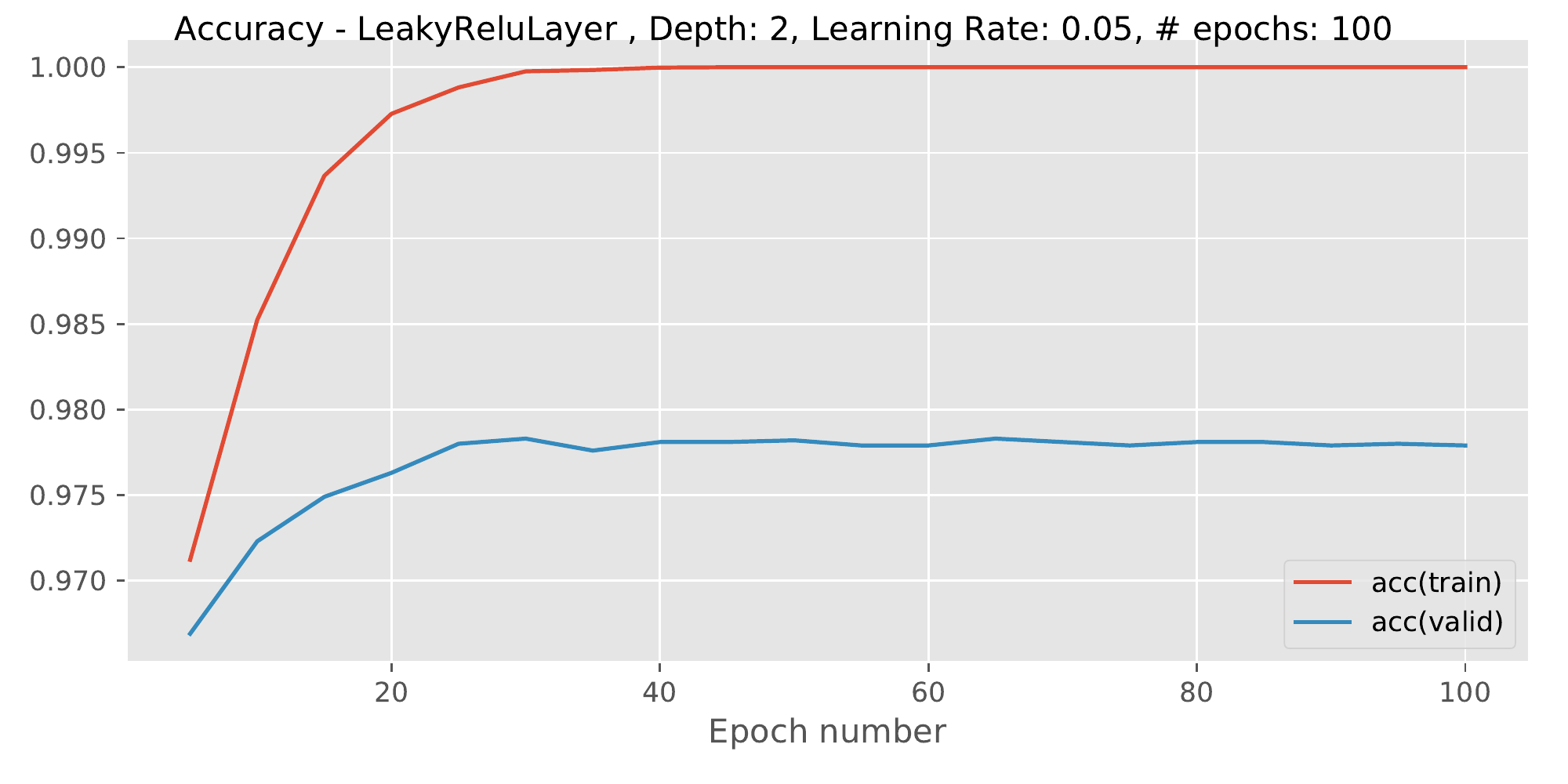}
}
\vskip -3mm
\caption{LeakyRelu - Error and accuracy with learning rate 0.05}
\label{fig:LeakyReluLayer_depth2_learningrate005_epochs100}
\end{center}
\end{figure}

\begin{figure}[H]
\vskip -3mm
\begin{center}
\subfigure{
\includegraphics[width=38mm, height=30mm]{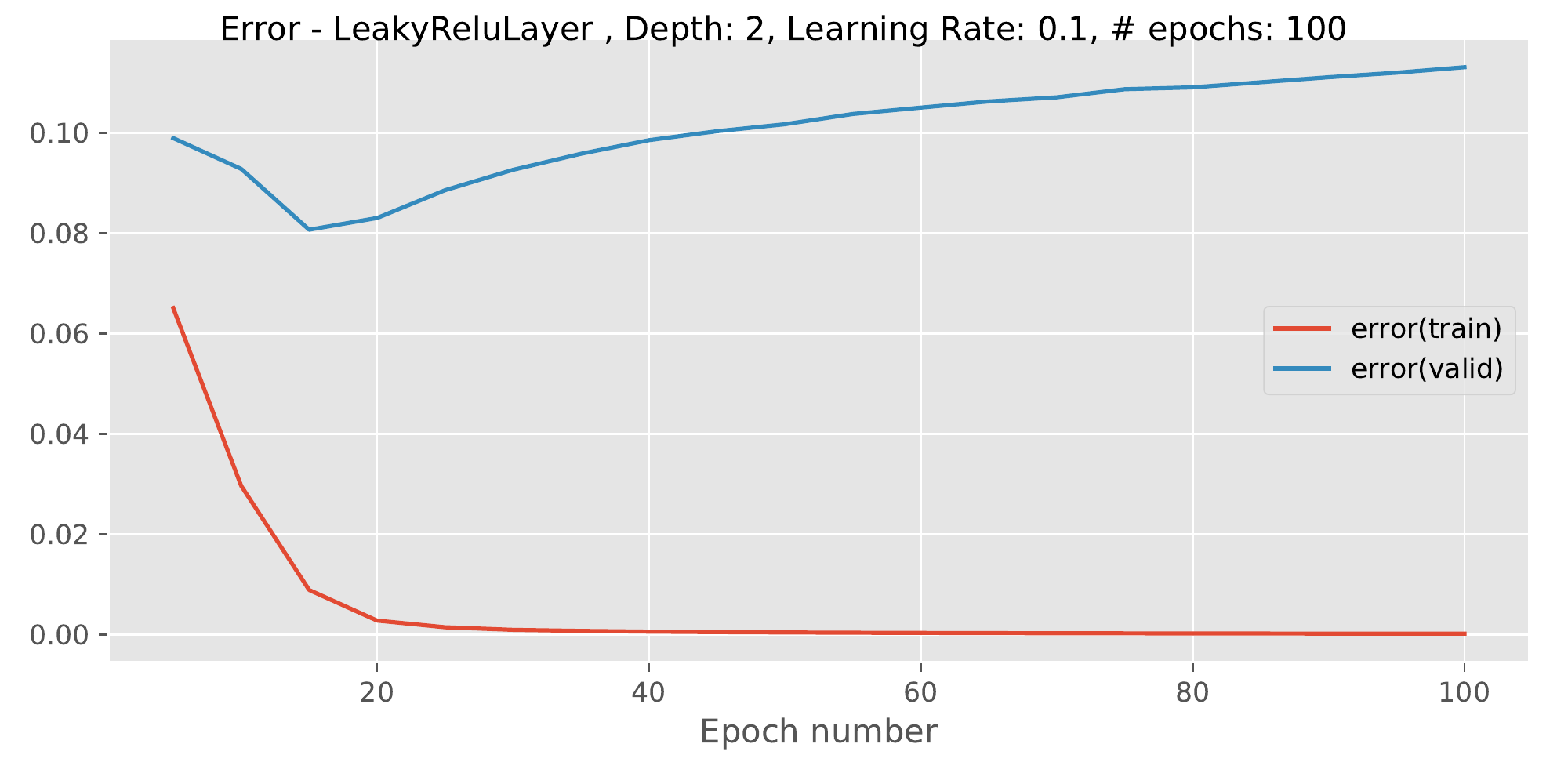}}
\subfigure{\includegraphics[width=38mm, height=30mm]{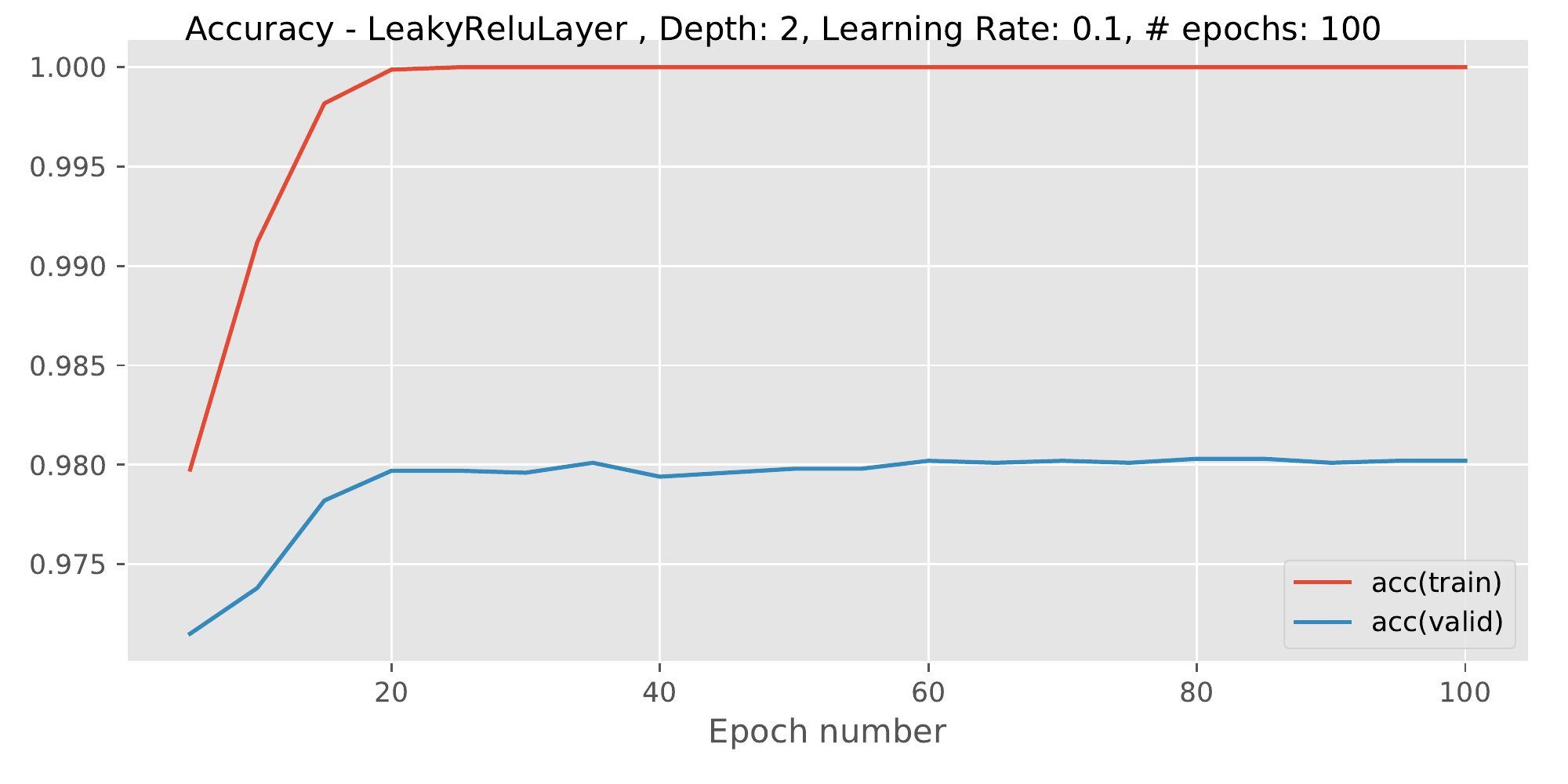}
}
\vskip -3mm
\caption{LeakyRelu - Error and accuracy with learning rate 0.10}
\label{fig:LeakyReluLayer_depth2_learningrate01_epochs100}
\end{center}
\end{figure}

\textbf{ELU:}
\begin{figure}[H]
\vskip -3mm
\begin{center}
\subfigure{
\includegraphics[width=38mm, height=30mm]{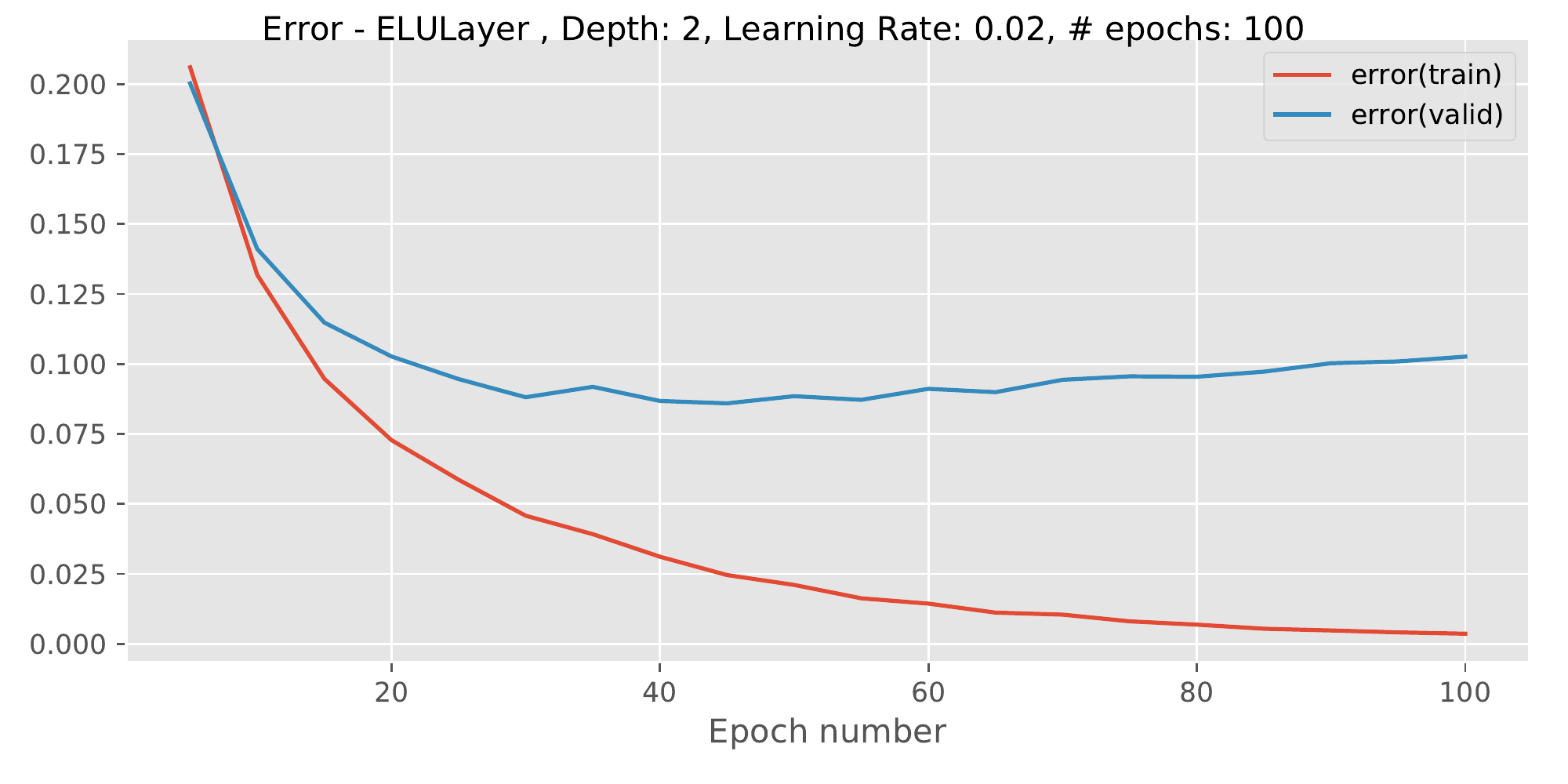}}
\subfigure{\includegraphics[width=38mm, height=30mm]{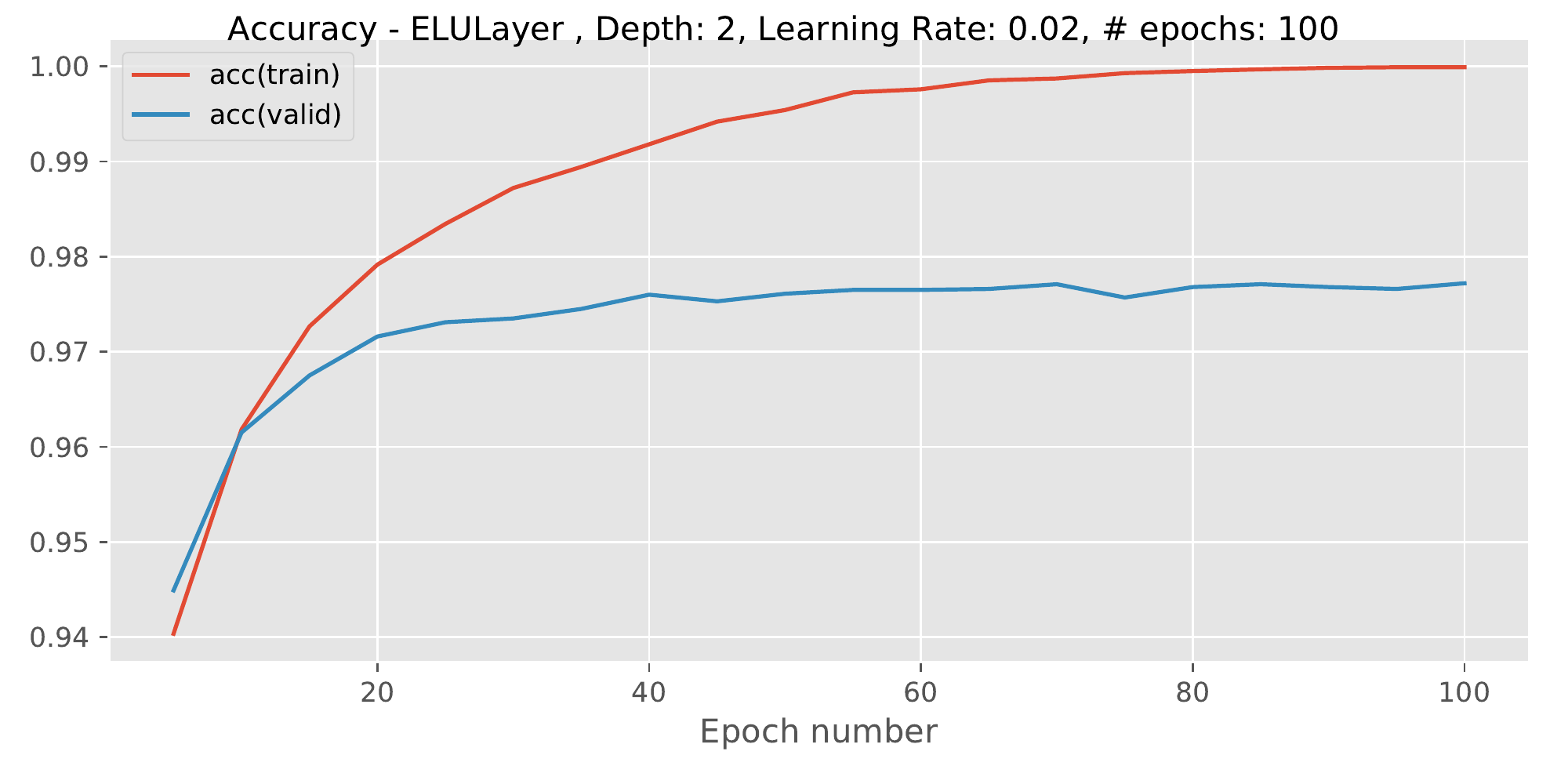}
}
\vskip -3mm
\caption{ELU - Error and accuracy with learning rate 0.02}
\label{fig:ELULayer_depth2_learningrate002_epochs100}
\end{center}
\end{figure}

\begin{figure}[H]
\vskip -3mm
\begin{center}
\subfigure{
\includegraphics[width=38mm, height=30mm]{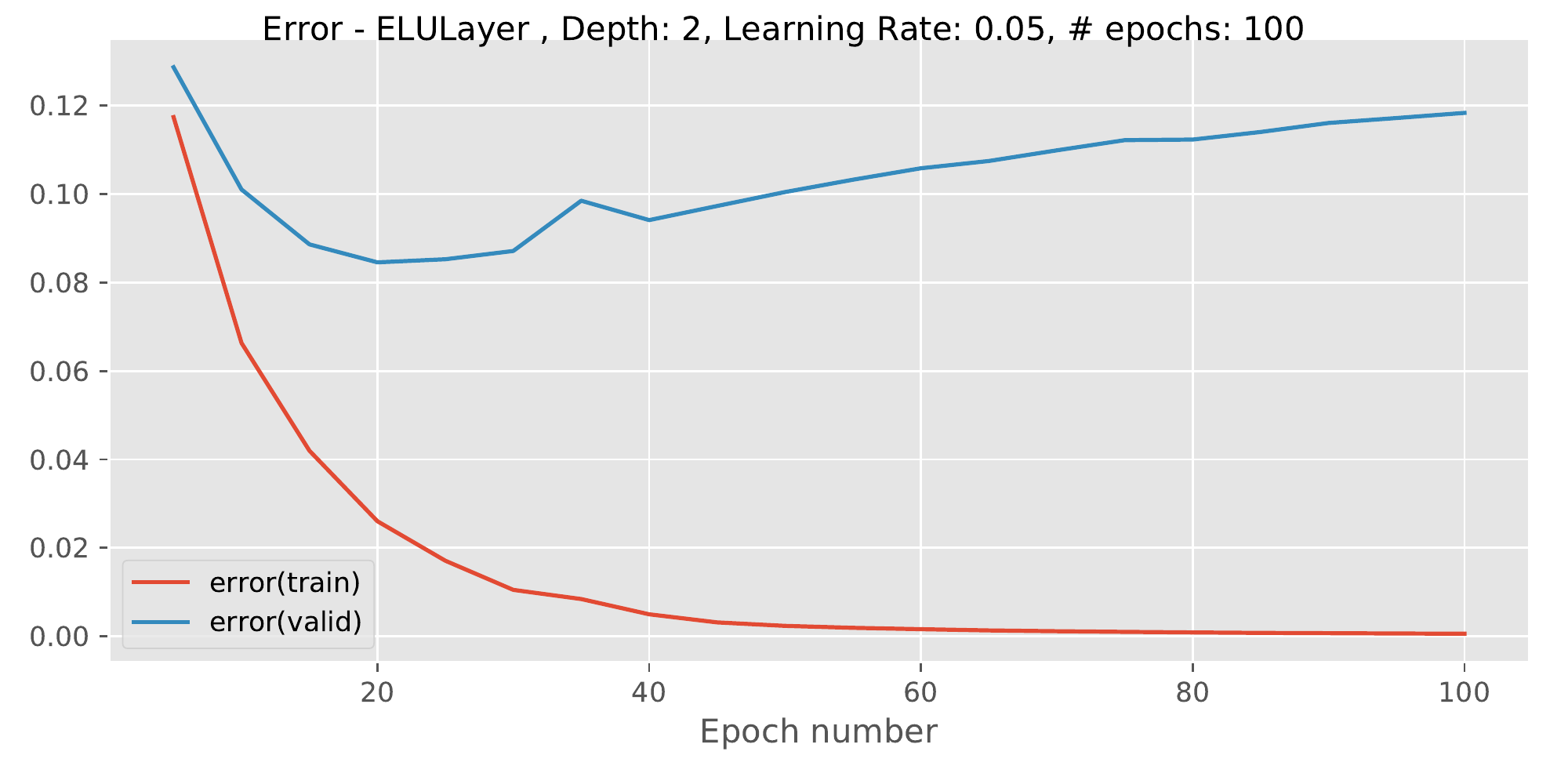}}
\subfigure{\includegraphics[width=38mm, height=30mm]{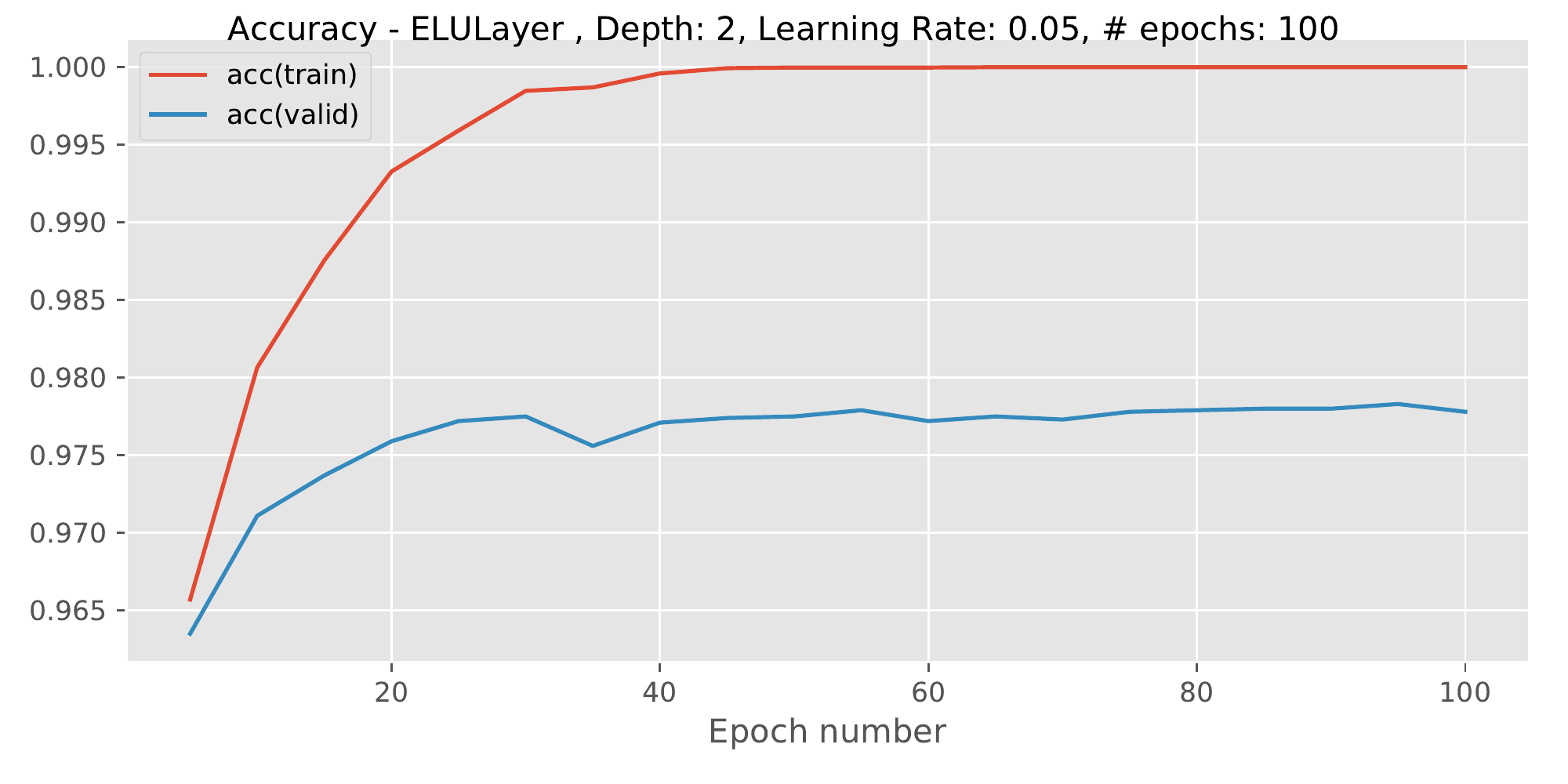}
}
\vskip -3mm
\caption{ELU - Error and accuracy with learning rate 0.05}
\label{fig:ELULayer_depth2_learningrate005_epochs100}
\end{center}
\end{figure}

\begin{figure}[H]
\vskip -3mm
\begin{center}
\subfigure{
\includegraphics[width=38mm, height=30mm]{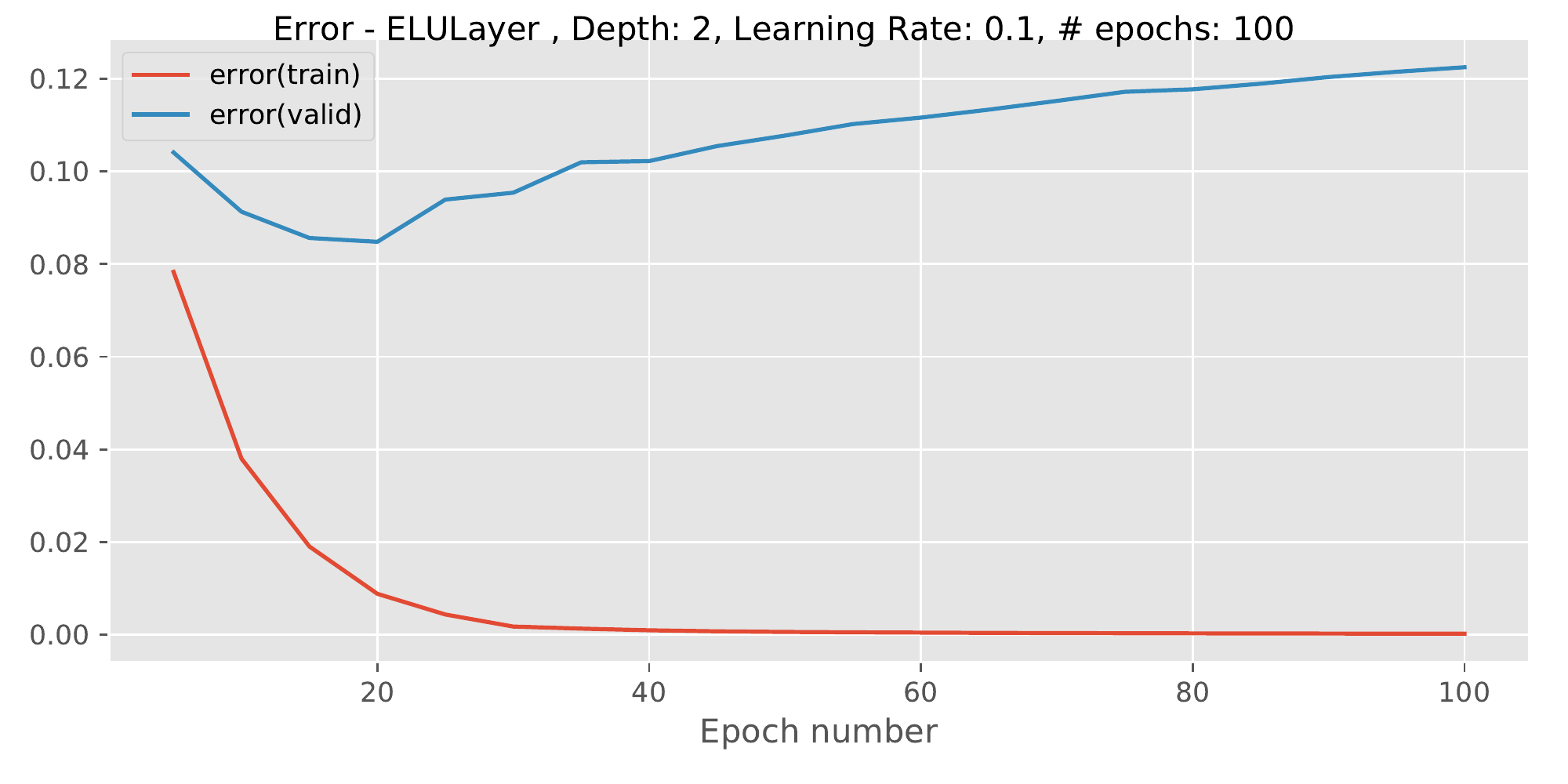}}
\subfigure{\includegraphics[width=38mm, height=30mm]{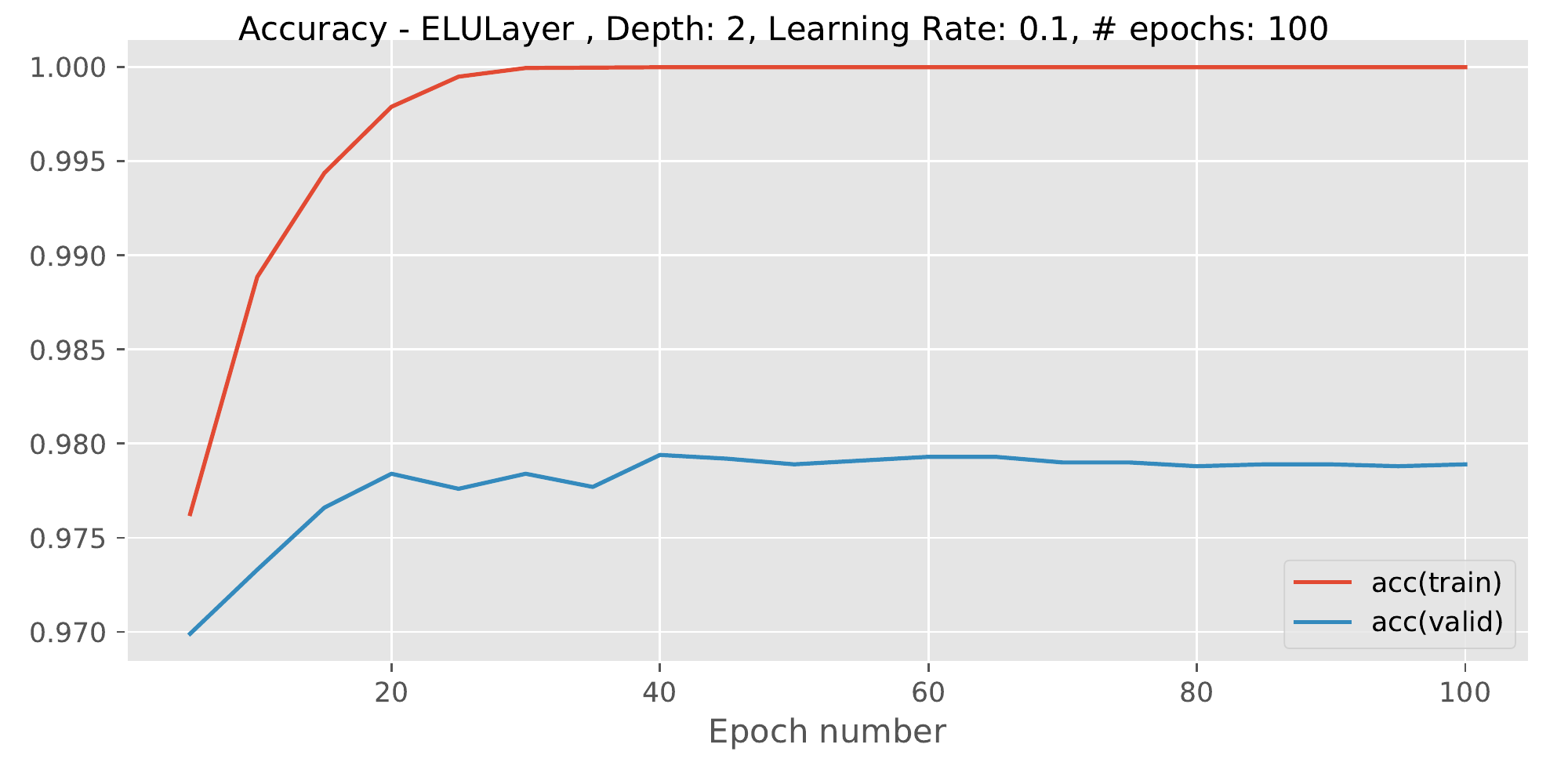}
}
\vskip -3mm
\caption{ELU - Error and accuracy with learning rate 0.10}
\label{fig:ELULayer_depth2_learningrate01_epochs100}
\end{center}
\end{figure}

\textbf{SELU:}
\begin{figure}[H]
\vskip -3mm
\begin{center}
\subfigure{
\includegraphics[width=38mm, height=30mm]{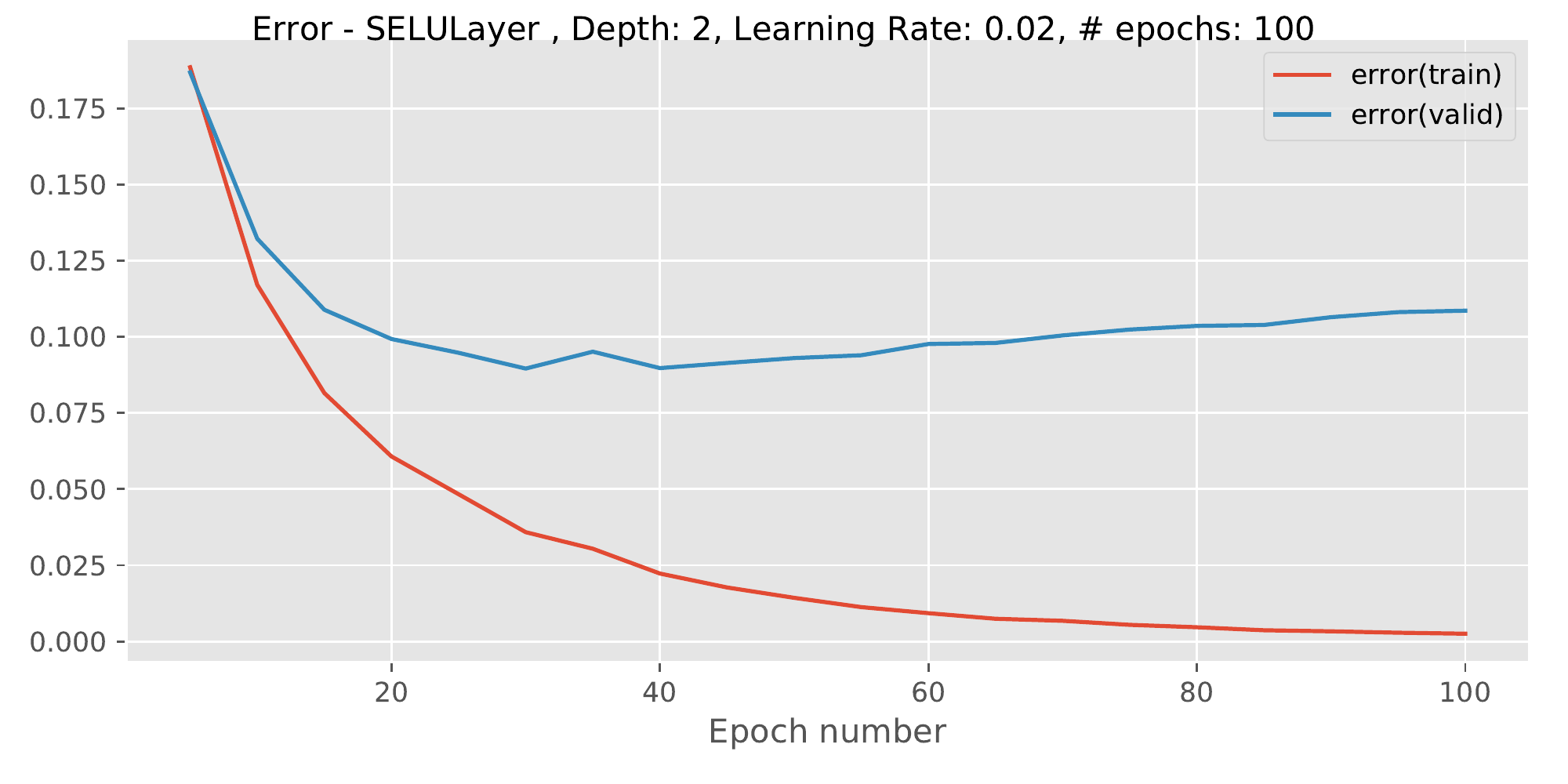}}
\subfigure{\includegraphics[width=38mm, height=30mm]{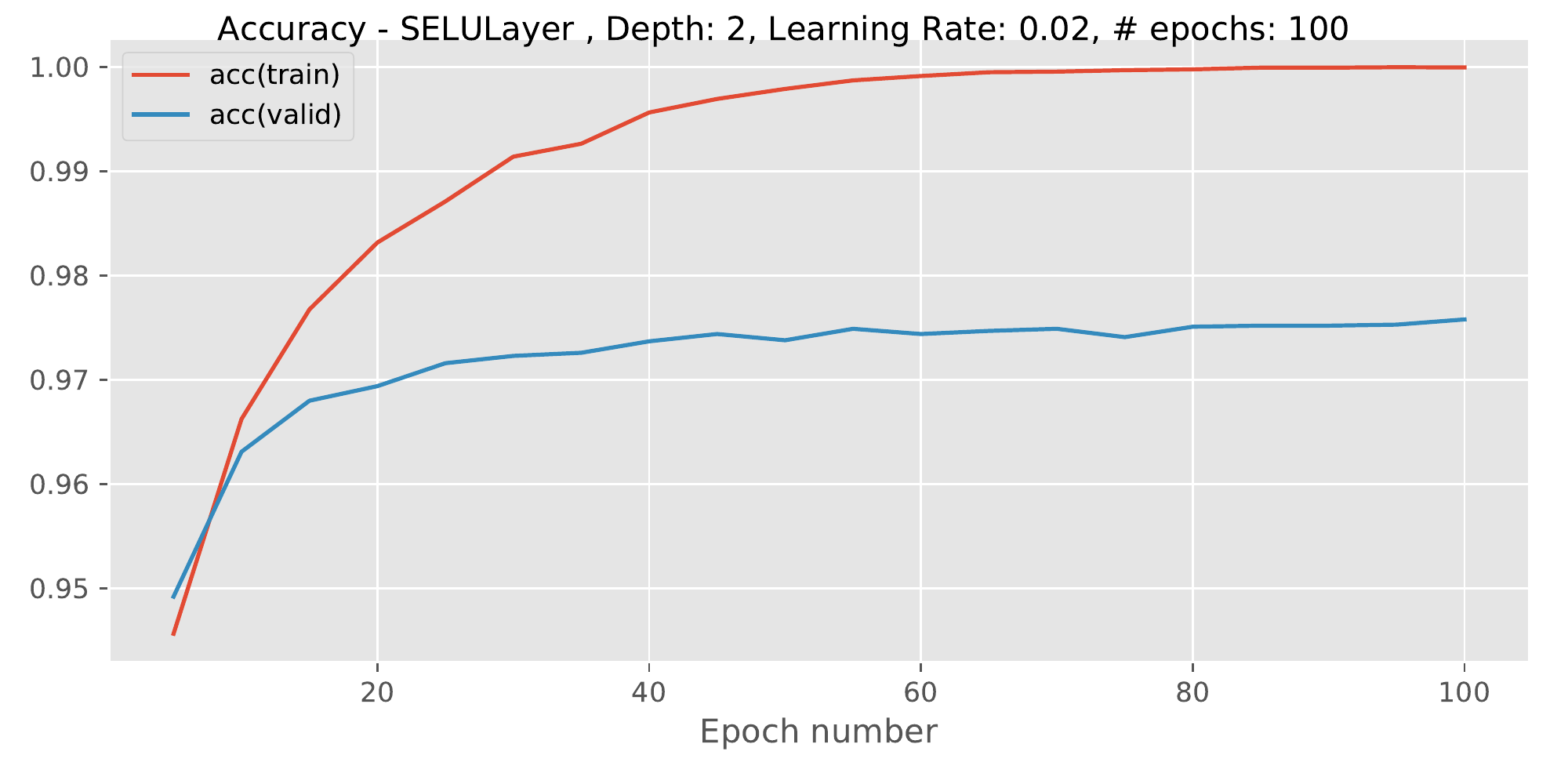}
}
\vskip -3mm
\caption{SELU - Error and accuracy with learning rate 0.02}
\label{fig:SELULayer_depth2_learningrate002_epochs100}
\end{center}
\end{figure}

\begin{figure}[H]
\vskip -3mm
\begin{center}
\subfigure{
\includegraphics[width=38mm, height=30mm]{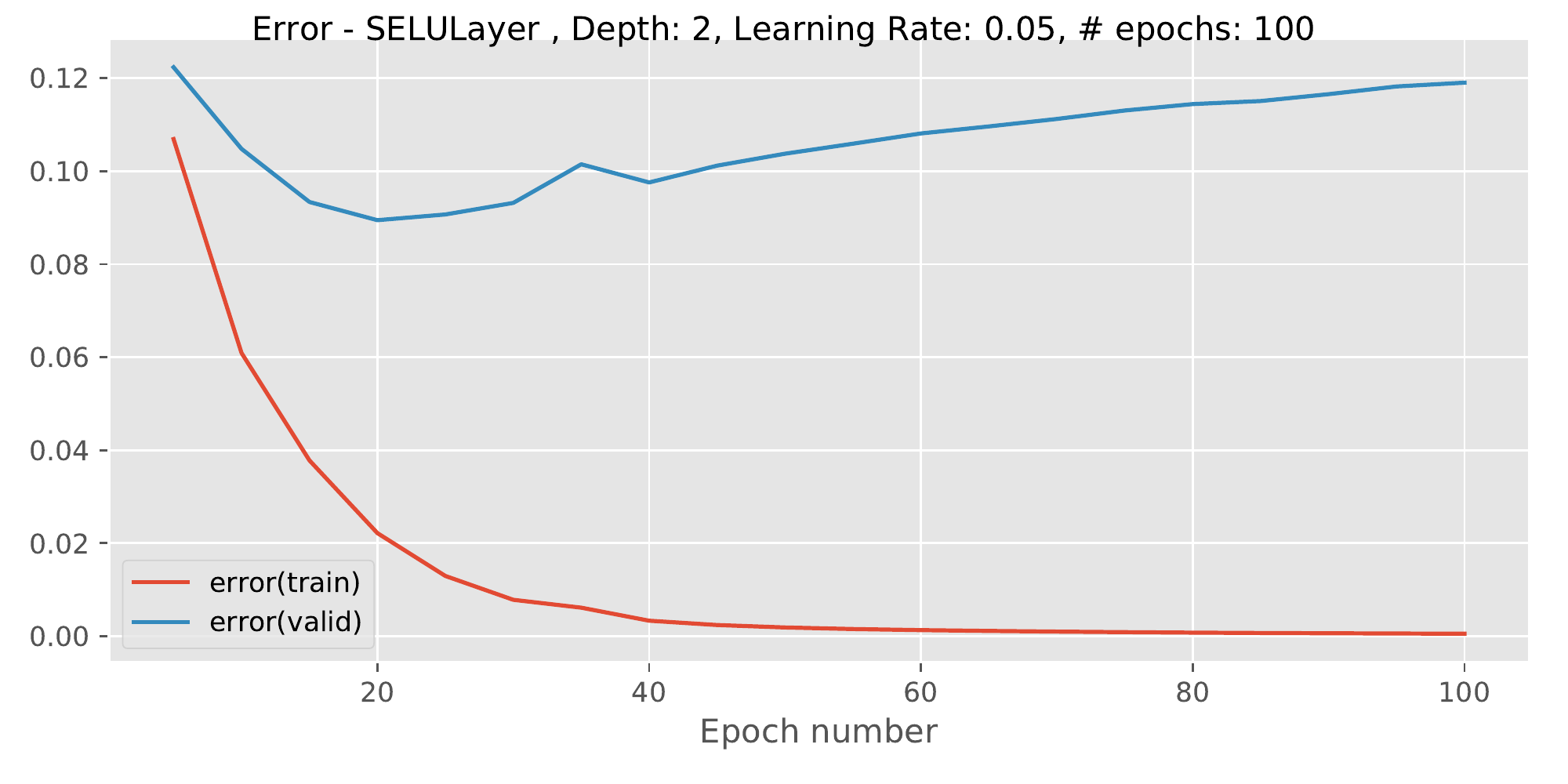}}
\subfigure{\includegraphics[width=38mm, height=30mm]{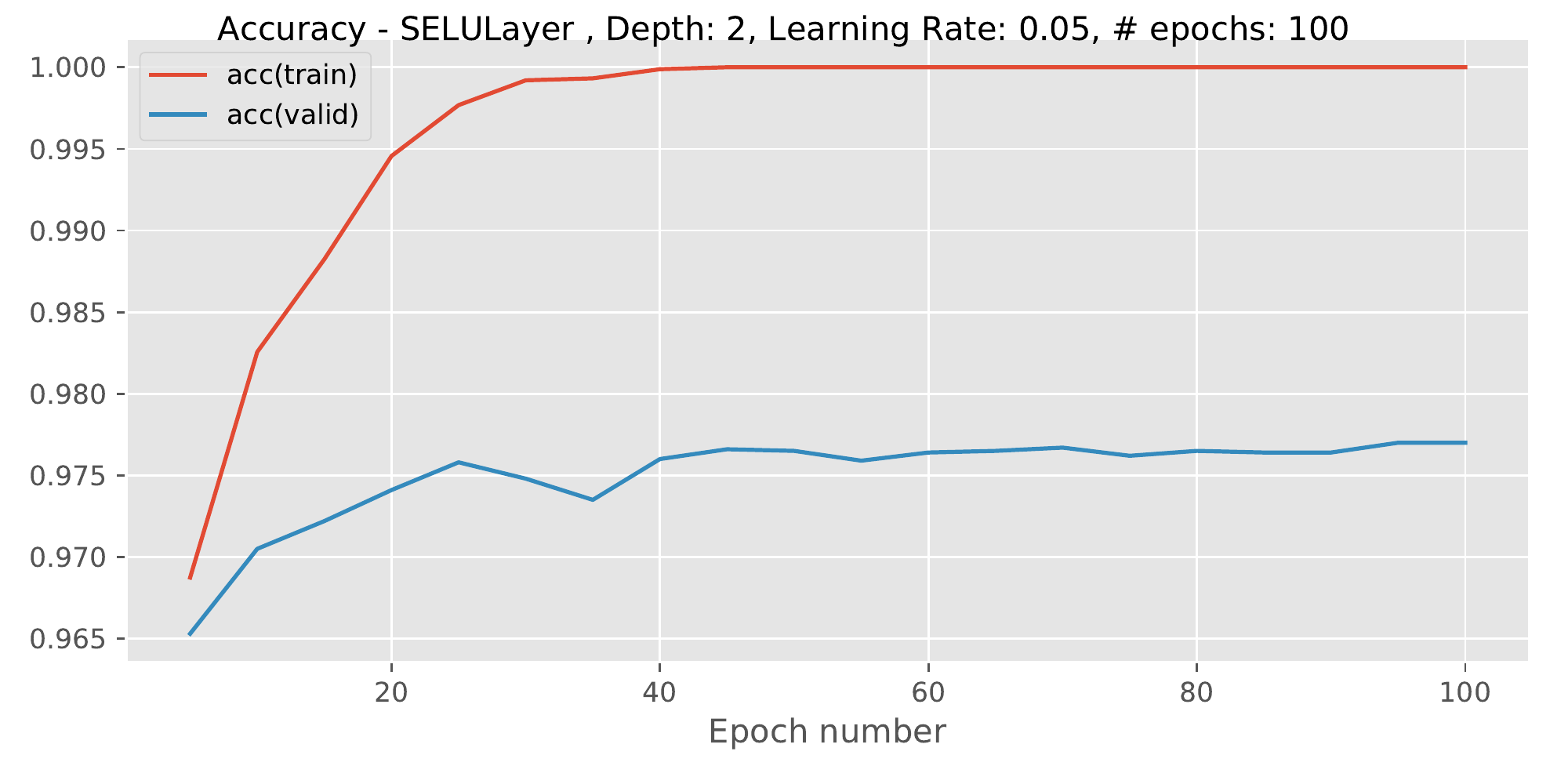}
}
\vskip -3mm
\caption{SELU - Error and accuracy with learning rate 0.05}
\label{fig:SELULayer_depth2_learningrate005_epochs100}
\end{center}
\end{figure}

\begin{figure}[H]
\vskip -3mm
\begin{center}
\subfigure{
\includegraphics[width=38mm, height=30mm]{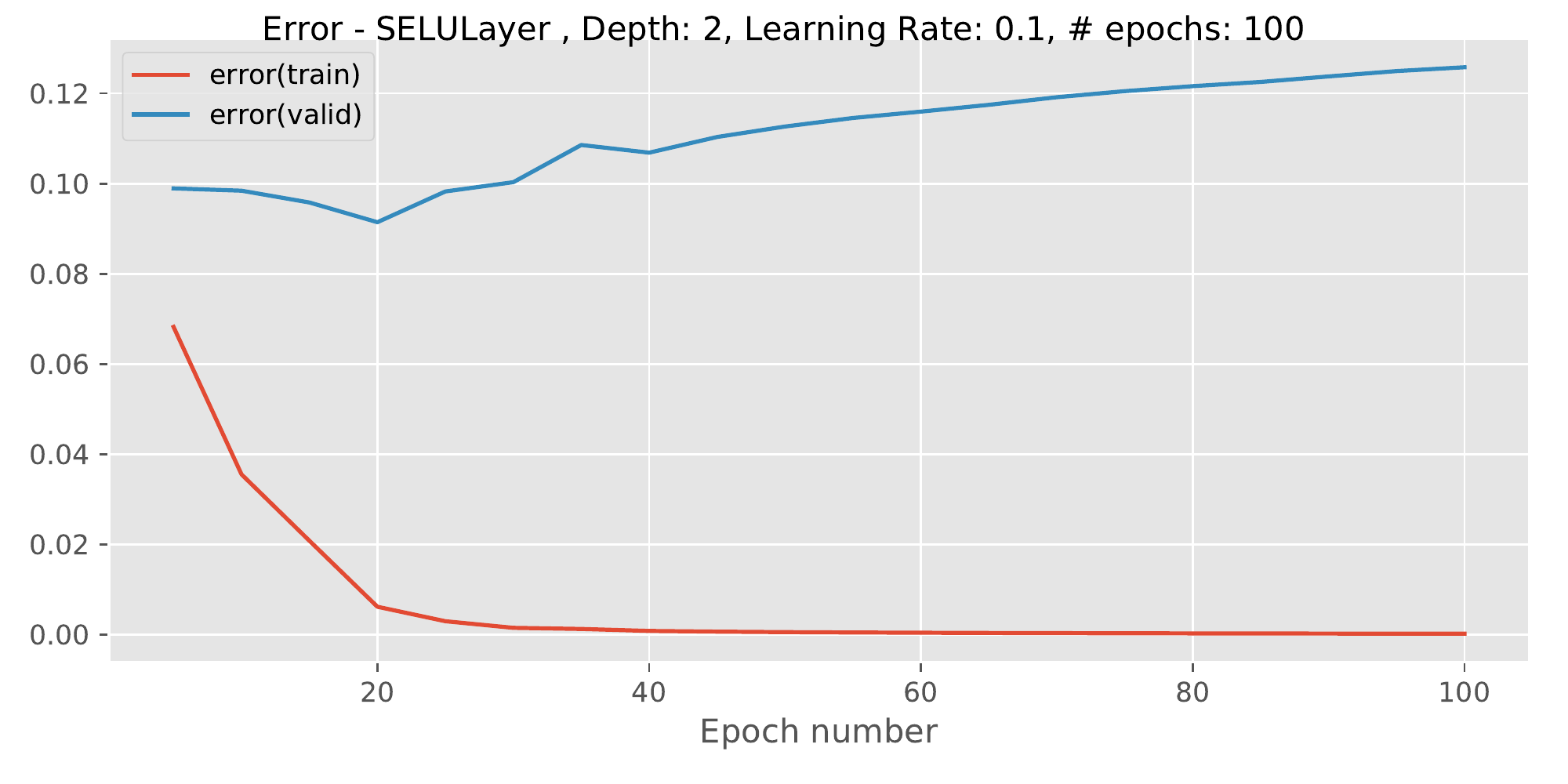}}
\subfigure{\includegraphics[width=38mm, height=30mm]{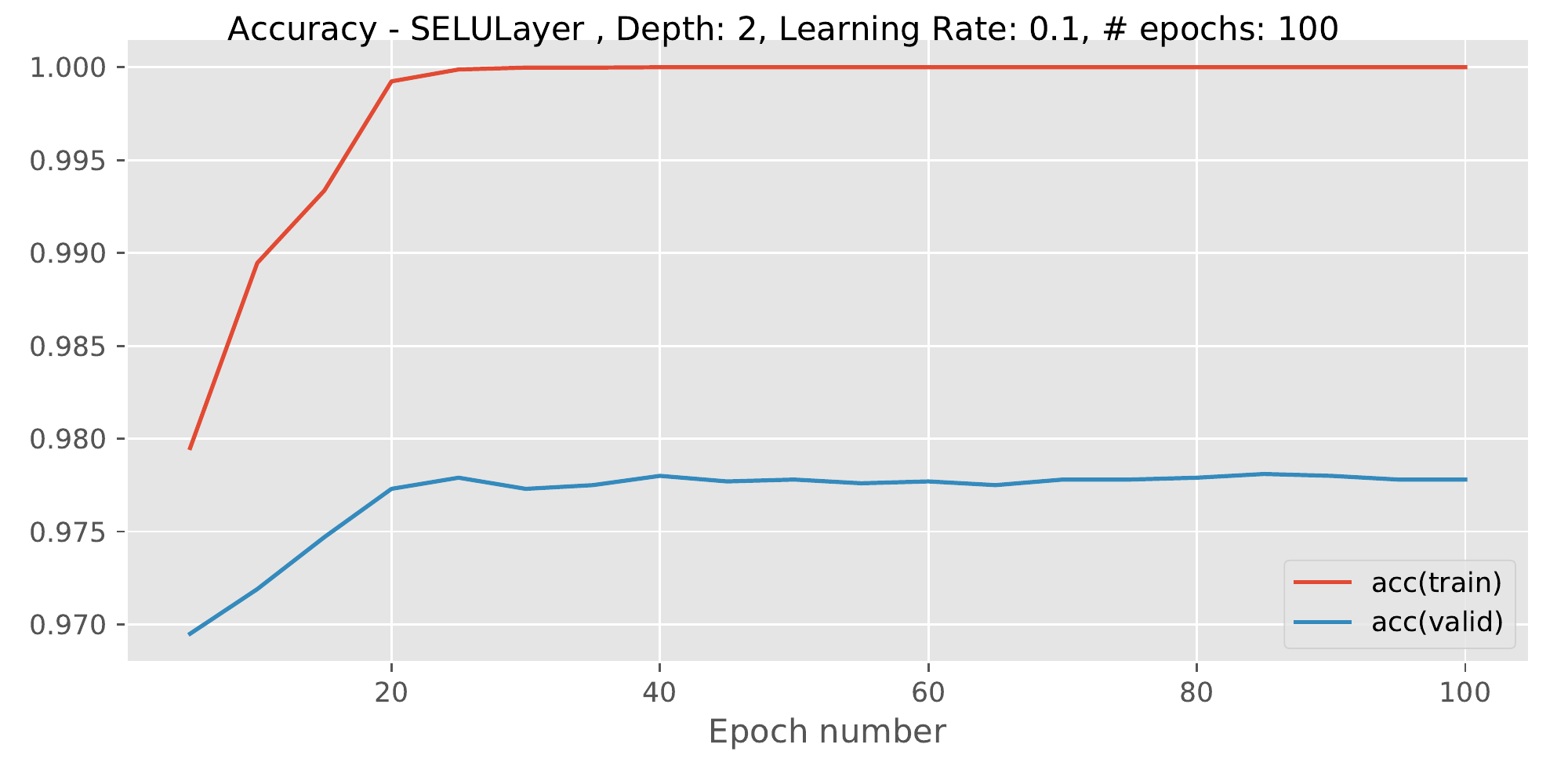}
}
\vskip -3mm
\caption{SELU - Error and accuracy with learning rate 0.10}
\label{fig:SELULayer_depth2_learningrate01_epochs100}
\end{center}
\end{figure}

\bibliography{DNNactivation2017}

\end{document}